\documentclass[10pt,twocolumn,letterpaper]{article}

\usepackage{cvpr}              % To produce the CAMERA-READY version
\usepackage{multirow}
\usepackage{marvosym}

\definecolor{cvprblue}{rgb}{0.21,0.49,0.74}
\usepackage[pagebackref,breaklinks,colorlinks,allcolors=cvprblue]{hyperref}
\usepackage{cuted}
\def\confName{CVPR}
\def\confYear{2026}

\title{HP-Edit: A Human-Preference Post-Training Framework for Image Editing}

\author{
Fan Li$^{1,3,*}$, Chonghuinan Wang$^{1,2,*}$, Lina Lei$^{1,3}$,
Yuping Qiu$^{1}$, Jiaqi Xu$^{1}$, Jiaxiu Jiang$^{1,2}$,\\ Xinran Qin$^{1}$, 
Zhikai Chen$^{1}$, Fenglong Song$^{1,}$\textsuperscript{\Letter}, Zhixin Wang$^{1}$, Renjing Pei$^{1, \dag}$, Wangmeng Zuo$^{2}$\\
{\small $^1$ Huawei Noah’s Ark Lab}
{\small $^2$ Harbin Institute of Technology}
{\small $^3$ Nankai University}
}
\begin{document}
\maketitle
\renewcommand{\thefootnote}{}
\footnotetext{\space * Equal Contribution, \space \dag \space Project Leader, \Letter \space Corresponding Author}
% \begin{figure*}[!h]
%     \centering
%     \includegraphics[width=\textwidth]{figure/teaser.png} 
%     \caption{Visual comparison of pretrained editing models before and after applying HP-Edit across eight common editing tasks.} 
%     \vspace{-0.3cm}
%     \label{fig:teaser}
% \end{figure*}
\begin{strip}
% \begin{figure*}
\centering
% \vspace{-40pt}
\includegraphics[width=\textwidth]{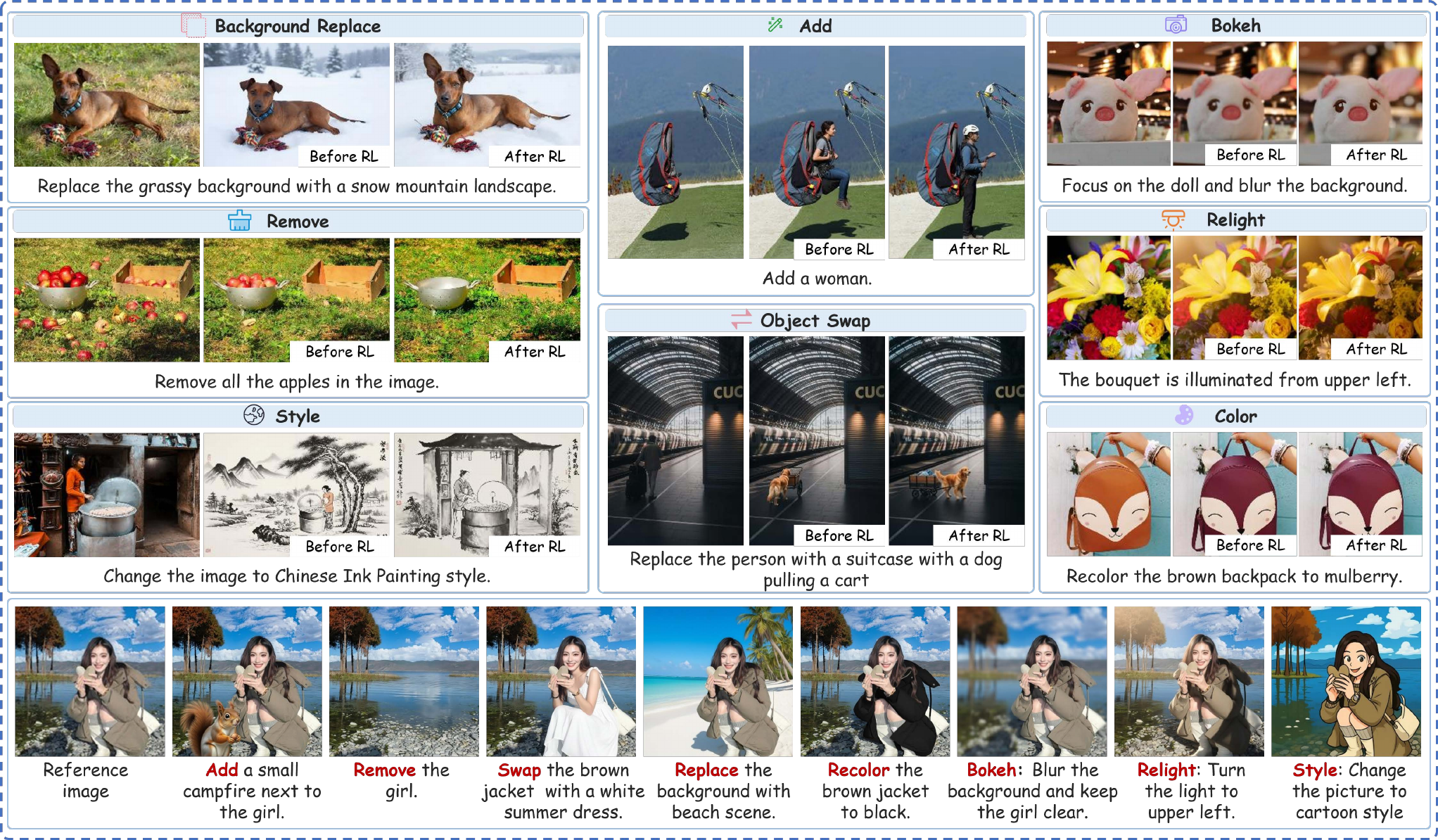} 
% \vspace{-7mm}
\captionof{figure}{
Visual comparison before and after applying HP-Edit based on the pretrained Qwen-Image-Edit-2509, across eight common editing tasks. We can clearly observe that the results after applying HP-Edit are more realistic and better aligned with human preferences.
}

% Replace the background with a serene forest scene.

% Replace the family of three with a group of three people dressed in summer attire, such as shorts and t-shirts, while keeping the background and setting unchanged.
% Erase all apples from the image.

% change this image to Japanese Ukiyo-e style, flat perspective, woodblock print texture, traditional Japanese colors, elegant composition, nature elements, cultural motifs, refined details, harmonious balance, ukiyo-e facial expressions, ukiyo-e landscape motifs

% Transform this image into the style of a Japanese Ukiyo-e woodblock print, featuring flat perspective, traditional Japanese colors, and characteristic elegant composition with refined details.

\label{fig:teaser}
\end{strip}
\begin{abstract}
% -v0- Recent reinforcement learning has become an important paradigm for text-to-image generation and image editing tasks, such as diffusion DPO and Flow-GRPO. However, efficiently applying Reinforcement Learning from Human Feedback (RLHF) to diffusion-based editing tasks remains largely unexplored. To address this limitation, we propose HP-Edit, a human-preference post-training framework for editing. HP-Edit applies a small amount of human preference scoring data and a Visual Large Language Model (VLLM) to quickly and efficiently construct a human preference dataset, which is then used to train the editing model to align with human-preferred outcomes. Additionally, this paper focuses on real-world editing and introduces the RealEdit-50K dataset, which balances common object editing while aligning with human preferences in real-world scenarios. After building the dataset, we use GRPO to train editing models based on a human-preference reward model. Extensive experiments demonstrate that our HP-Edit framework efficiently improves editing models like FLUX.1-Kontext and Qwen-Image-Edit, enhancing image editing with specific human preferences.

Common image editing tasks typically adopt powerful generative diffusion models as the leading paradigm for real-world content editing. Meanwhile, although reinforcement learning (RL) methods such as Diffusion-DPO and Flow-GRPO have further improved generation quality, efficiently applying Reinforcement Learning from Human Feedback (RLHF) to diffusion-based editing remains largely unexplored, due to a lack of scalable human-preference datasets and frameworks tailored to diverse editing needs.
To fill this gap, we propose HP-Edit, a post-training framework for \textbf{H}uman \textbf{P}reference-aligned \textbf{Edit}ing, and introduce RealPref-50K, a real-world dataset across eight common tasks and balancing common object editing.
Specifically, HP-Edit leverages a small amount of human-preference scoring data and a pretrained visual large language model (VLM) to develop HP-Scorer—an automatic, human preference-aligned evaluator. We then use HP-Scorer both to efficiently build a scalable preference dataset and to serve as the reward function for post-training the editing model.
We also introduce RealPref-Bench, a benchmark for evaluating real-world editing performance. Extensive experiments demonstrate that our approach significantly enhances models such as Qwen-Image-Edit-2509, aligning their outputs more closely with human preference.
\end{abstract}    
\section{Introduction}
\label{sec:intro}

Text-to-image (T2I) generation and image-to-image (I2I) editing have become foundational technologies for content creation across industries, ranging from digital design and product marketing to real-world scene customization. Diffusion models have emerged as the de facto standard due to their high-quality, controllable outputs~\cite{ho2020denoising, Rombach_2022_LDM, FLUX, SD2}. For I2I editing, state-of-the-art models~\cite{labs2025flux, wu2025qwen} typically build on pretrained T2I backbones via Supervised Fine-Tuning (SFT), leveraging large-scale I2I datasets to acquire editing capabilities. However, two critical limitations plague SFT-based approaches: first, the mixed sources of SFT data (e.g., cartoons, synthetic images) often misalign with real-world human preferences; second, constructing preference-aligned editing datasets requires expensive manual annotation, making scalable alignment impractical.

Reinforcement learning (RL)~\cite{sutton1998reinforcement} has proven to be highly effective in enhancing the reasoning and alignment capabilities of large language models (LLMs)~\cite{guo2025deepseek, jaech2024openai}, and recent works such as Diffusion-DPO~\cite{wallace2024diffusion}, Flow-GRPO~\cite{liu2025flow}, and Dance-GRPO~\cite{xue2025dancegrpo} have shown promise in improving the quality of T2I generation through post-training. Despite this progress, RL-driven human-preference alignment for I2I editing remains underexplored. Unlike open-ended T2I synthesis, I2I editing demands both task accuracy (e.g., faithfully removing an object) and preference alignment (e.g., natural-looking results). This dual objective calls for frameworks that combine efficient preference data construction—without prohibitive annotation costs—with task-aware reward models tailored to diverse editing sub-tasks. Furthermore, existing editing research lacks a real-world, object-balanced benchmark, hindering accurate evaluation of preference-aligned editing.

Our work addresses these gaps with three key contributions. First, we present HP-Edit, a post-training framework for \textbf{H}uman \textbf{P}reference \textbf{Edit}ing that unifies a Visual Large Language Model (VLM)-based scorer, HP-Scorer, an efficient hard-case-focused dataset construction pipeline, and task-aware RL post-training to align models with human preferences while preserving editing accuracy. Second, we introduce RealPref-50K, a real-world-oriented dataset of over 50K cases across eight common sub-tasks—addition, removal, background replacement, object swapping, color change, bokeh (background defocus), relighting, and style transfer—covering common MS-COCO~\cite{lin2014microsoft} objects (e.g., \textit{person, car, cake}) with a balanced distribution to better reflect practical needs. Third, we release RealPref-bench, a benchmark for evaluating preference-aligned editing using real-world images and manually verified preference instructions, enabling rigorous model evaluation.

In HP-Edit, we first collect a small number of triples (input image, edited output, instruction) per editing sub-task and annotate each triple with human-preference scores ranging from 0 to 5 (see the supplementary material for the scoring criteria). The proposed HP-Scorer is built on a pretrained VLM (e.g., Qwen2.5-VL~\cite{bai2025qwen2}) with a progressively optimized task-aware scoring prompt to approximate human judgments. We then run inference on existing editing datasets using the pretrained editing model and use the VLM-based HP-Scorer to filter hard cases that best capture preference signals, constructing a scalable RL training dataset. Finally, we employ the HP-Scorer as a task-aware reward model and apply RL post-training to the pretrained editing model. Extensive experiments on FLUX.1-Kontext-dev and Qwen-Image-Edit-2509 demonstrate that HP-Edit achieves significant improvements in both human-preference ratings and visual quality, compared to strong pretrained baselines.

% The rest of this paper is organized as follows. Section~2 reviews related work on diffusion-based generation and editing, and RL for T2I. Section~3 presents preliminaries on flow matching and Flow-GRPO. Section~4 introduces the HP-Edit framework as well as the RealPref-50K dataset and the RealPref-Bench benchmark. Section~5 reports the experimental setup and results. Section~6 discusses limitations and directions for future work.

\begin{figure*}[t]
    \centering
    \includegraphics[width=\textwidth]{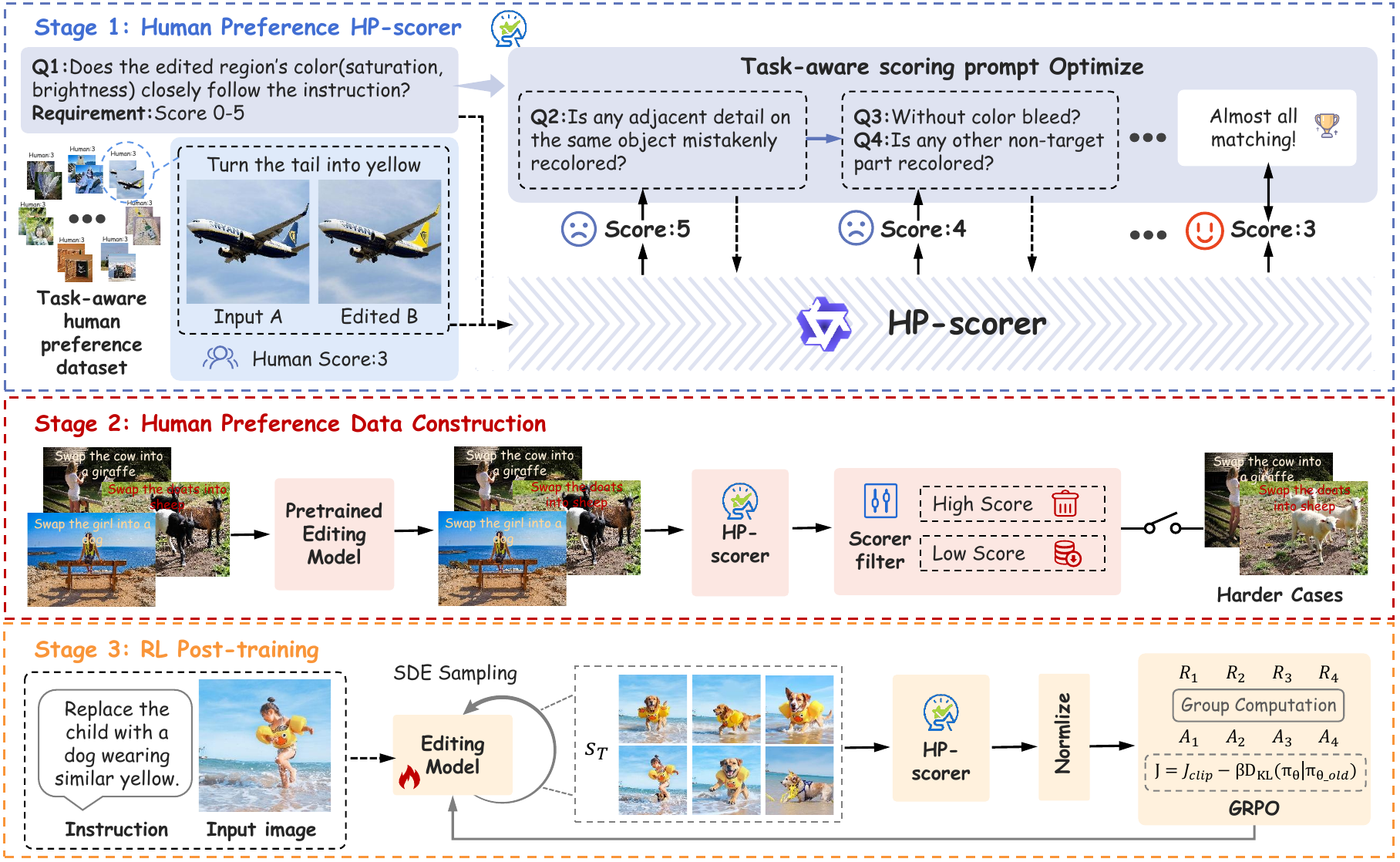} 
    \caption{The overview of the proposed framework, HP-Edit which consists of three stages:  the task-aware HP-Scorer for human preference scoring human preference, the data pipeline of human preference data and the human preference RL post-training.} 
    \vspace{-0.3cm}
    \label{fig:pipeline}
\end{figure*}

\section{Related Work}
\label{sec:relatedwork}

\subsection{Diffusion Models}
In recent years, diffusion models \cite{sohl2015deep, ho2020denoising, song2020denoising, song2020score, nichol2021improved} have achieved remarkable progress in generative tasks. Early diffusion models are driven by a stochastic differential equation (SDE) and optimize denoising-based score matching objectives \cite{ho2020denoising,song2020score,song2020denoising}, commonly minimizing the MSE between the predicted noise and the ground-truth perturbation. Recently, Rectified Flow \cite{liu2022flow} and Flow Matching \cite{lipman2022flow} adopt a deterministic probability flow ODE to reduce dependence on noise assumptions and improve stability and scalability. Diffusion models have expanded beyond generic image synthesis to encompass text-to-image and video generation \cite{li2024brushedit,huang2024smartedit,shi2024seededit,yu2025anyedit,kong2025dual,li2024magiceraser, qin2025camedit,sun2025pocketsr,wang2025ace,wu2025vtinker}, as well as a wide range of image restoration and editing.
%, directly regressing the velocity field to match model trajectories with the target probability flow, thereby
\subsection{Image Editing}
Research on image editing centers on controllability, local fidelity, and tight coupling with strong base generators. Data-driven fine-tuning and scalable pipelines\cite{brooks2023instructpix2pix,zhang2023magicbrush, li2024brushedit} improve editing realism, while architectural and adaptation techniques \cite{huang2024smartedit,shi2024seededit,yu2025anyedit,zhao2024ultraedit} enhance fine-grained control and efficiency. Unified frameworks \cite{xiao2025omnigen,wu2025omnigen2,han2024ace,mao2025ace++,liu2025step1x,deng2025Bagel,ma2025x2edit,lin2025uniworld} merge generation and editing into a unified stack. FLUX.1-Kontext \cite{labs2025flux} adds high-consistency reference conditioning for rapid, reference-guided edits. Qwen-Image \cite{wu2025qwen} leverages large-scale curriculum training for precise edits and complex text rendering.

\subsection{Learning from Human Feedback}
In generative vision, aligning diffusion or flow models with human preferences has shifted from reliance on automatic quantitative proxies to explicit preference-based optimization. Direct Preference Objectives (DPO) \cite{rafailov2023direct} improves human-preference alignment by increasing the log-probability of preferred samples. Subsequent works refine this paradigm with robust weighting and score-matching-based formulations \cite{wallace2024diffusion,tamboli2025balanceddpo,zhu2025dspo}, and extend it beyond single-image to multi-image, video, and editing tasks \cite{liu2024mia,liu2025videodpo,wu2025densedpo,hu2025d}.
In parallel, the integration of Group Relative Policy Optimization (GRPO) \cite{shao2024deepseekmath} into modern flow models converts deterministic probability-flow ODEs into marginally consistent SDEs via efficiency-aware heuristics, enabling few-step online preference alignment \cite{lipman2022flow,liu2022flow,liu2025flow}. 
Subsequent GRPO variants expand applicability to diffusion/flow formulations, supporting text-to-image, text-to-video, and image-to-video tasks \cite{xue2025dancegrpo,he2025tempflow,wang2025pref,luo2025sample}.
\section{Preliminaries}

\subsection{Flow Matching}
Flow matching \cite{lipman2022flow,liu2022flow} serves as a generative modeling paradigm for training continuous normalizing flows by aligning modeled velocity fields with those derived from data interpolations. In the rectified flow \cite{liu2022flow} formulation, a sample $\mathbf{x}_t$ at time $t \in [0,1]$ is defined as a linear interpolation between a base sample $\mathbf{x}_0 \sim p_0$ (typically Gaussian) and a data sample $\mathbf{x}_1 \sim p_1$:
\begin{equation}
    \mathbf{x}_t = (1 - t)\mathbf{x}_0 + t \mathbf{x}_1.
\end{equation}
The model learns a neural velocity field $\mathbf{v}_\theta(\mathbf{x}_t, t)$ to approximate the target velocity field:
\begin{equation}
    \mathcal{L}_{\text{FM}} = \mathbb{E}_{t, \mathbf{x}_0, \mathbf{x}_1} \left[ \left\| \mathbf{v}_\theta(\mathbf{x}_t, t) - (\mathbf{x}_1 - \mathbf{x}_0) \right\|^2 \right],
\end{equation}
where the corresponding target velocity field is the constant vector from $\mathbf{x}_0$ to $\mathbf{x}_1$.

\subsection{Flow-GRPO}

As established in \cite{black2023training}, the iterative reverse-time sampling process of a flow matching model can be formulated as a Markov Decision Process (MDP), defined by the tuple $(\mathcal{S}, \mathcal{A}, \rho_0, \mathcal{P}, R)$.
For a given prompt $c$, the policy $\pi_\theta$ (the flow model) generates a trajectory $(s_0,a_0,s_1,a_1,...,s_T)$, where $\pi_\theta(a_t | s_t) \triangleq p_\theta(x_{t-1} | x_t, c)$.
The reward is provided only at the final step $t=T$ by the reward model $R=R(x_T, c)$, which quantifies the quality of $\mathbf{x}_T$ or its alignment with the prompt $\mathbf{c}$.

Group Relative Policy Optimization (GRPO) \cite{shao2024deepseekmath} is a lightweight and memory-efficient online reinforcement learning algorithm.
For flow matching, GRPO computes advantages by comparing rewards within a group of samples generated from the same prompt.
Specifically, for a given prompt $c$, the policy $\pi_\theta$ generates a group of $G$ images, resulting in a set of final images $\{x_T^i\}_{i=1}^G$.
The advantage $\hat{A}^i$ for each sample in the group is then calculated by normalizing its reward relative to the group's statistics:
\begin{equation}
    \label{eq:grpo_advantage}
    \hat{A}^i = \frac{R(x_T^i, c) - \text{mean}\left(\{R(x_T^j, c)\}_{j=1}^G\right)}{\text{std}\left(\{R(x_T^j, c)\}_{j=1}^G\right)}
\end{equation}

The policy model's parameters $\theta$ are subsequently updated by maximizing the GRPO objective function, which encourages actions leading to above-average rewards while ensuring training stability. The objective $J(\theta)$ is defined as an expectation over trajectories sampled from the previous policy $\pi_{\theta_{\text{old}}}$:
\begin{equation}
\label{eq:grpo_objective_full}
\begin{aligned}
    & J_{\text{Flow-GRPO}}(\theta) = \mathbb{E}_{c \sim \mathcal{C}, \{x^i\} \sim \pi_{\theta_{\text{old}}}(\cdot|c)} \Biggl[ \frac{1}{G} \sum_{i=1}^{G} \frac{1}{T} \sum_{t=0}^{T-1} \biggl( \\
    & \text{min} \left( r_t^i(\theta) \hat{A}^i, \text{clip}(r_t^i(\theta),1-\epsilon,1+\epsilon) \hat{A}^i \right) 
    \\ & - \beta D_{\text{KL}}\left(\pi_\theta \| \pi_{\text{ref}} \right) \biggr) \Biggr],
    \ \text{where} \ r_t^i(\theta) = \frac{p_\theta(x_{t-1} | x_t,c)}{p_{\theta_{old}}(x_{t-1} | x_t,c)}.
\end{aligned}
\end{equation}

GRPO relies on stochastic policy sampling for exploration, whereas the generative process in standard Flow Matching is governed by a deterministic ODE:
\begin{equation}
    d\mathbf{x}_t = \mathbf{v}_t dt.
\end{equation}

To address this, Flow-GRPO converts the deterministic ODE into an equivalent SDE, where the marginal probability density of the SDE at any time $t$ is guaranteed to match that of the original ODE flow:
\begin{equation}
    d\mathbf{x}_t = \left( \mathbf{v}_t(\mathbf{x}_t, t) + \frac{\sigma_t^2}{2t} (\mathbf{x}_t + (1-t)\mathbf{v}_t(\mathbf{x}_t, t)) \right) dt + \sigma_t d\mathbf{w},
\end{equation}
where $\mathbf{w}$ is a standard Wiener process and $\sigma_t$ controls the magnitude of the stochasticity during generation.
\section{Approach}
In this work, as shown in Figure~\ref{fig:pipeline}, we propose HP-Edit, a post-training framework for human preference-aligned editing, and introduce RealPref-50K, a real-world dataset that balances common object editing with human preferences. We also construct RealPref-Bench, an editing benchmark to effectively evaluate real-world editing performance.

\subsection{Overview of the Framework}
\label{sec:framework}

Although RL-based post-training techniques provide a suitable paradigm for human preference–aligned image editing, a key challenge remains: developing a post-training framework that integrates an efficient data construction pipeline for online training and a high-quality, task-specific reward model tailored to diverse editing tasks, thus fully leveraging the pretrained model’s capabilities while aligning with human preferences. To address this challenge, we propose \textbf{HP-Edit} that comprises three key stages: 
(1) optimization of the \textbf{H}uman \textbf{P}reference–scorer (HP-\textbf{S}corer); 
(2) an efficient dataset construction pipeline focusing on hard cases for preference learning guided by the HP-Scorer; and 
(3) a task-aware post-training stage where the HP-Scorer serves as the reward function.

Before detailing the framework, we first clarify the $0$–$5$ scoring criteria used for both human annotators and the visual language model (VLM). Given an editing triple consisting of ``input image~A'', ``edited image~B'', and an ``instruction'', the scoring guideline is as follows:

\begin{itemize}
    \item \textbf{Score 0:} The edited image~B is completely incorrect, does not follow the instruction at all, or fails to meet any requirements.
    \item \textbf{Score 1:} The edited image~B is partially correct but still largely incorrect. It follows the instruction only marginally, or the result appears unrealistic.
    \item \textbf{Score 2:} The edited image~B is mostly correct but still deviates from the instruction or fails to meet several key requirements.
    \item \textbf{Score 3:} The edited image~B generally follows the instruction, but its visual quality or aesthetics are subpar.
    \item \textbf{Score 4:} The edited image~B largely follows the instruction, and the visual quality and aesthetics are good.
    \item \textbf{Score 5:} The edited image~B fully follows the instruction, satisfies all requirements, and exhibits high-quality, realistic visual results.
\end{itemize}

\textbf{HP-Scorer.} As shown in Stage~1 of Figure~\ref{fig:pipeline}, we first collect a small number of editing cases (input image~A, instruction~T)—approximately 50–100 per editing sub-task—and apply the pretrained editing model to generate the edited image~B, forming editing triples. These triples are then manually rated by human annotators using the $0$–$5$ scoring scale.
Using these annotated samples, we employ a pretrained visual language model (VLM), such as Qwen3-VL and GPT-4o, as HP-Scorer with a carefully designed scoring system prompt for each sub-task. Each sub-task is assigned a carefully designed scoring prompt. The process begins with a basic scoring prompt containing only the aforementioned criteria and is iteratively refined by adding task-specific reasoning questions (e.g., for the \textit{object swapping} task: ``Is the object replacement feasible and clearly specified?'', ``Is the original object completely replaced?''). This refinement continues until the HP-Scorer's scoring results closely match human judgments across the collected triples. Notably, the score generated by the HP-Scorer is directly adopted as the \textbf{HP-score} used for evaluation.

\textbf{Human preference data construction pipeline.} As shown in Stage~2 of Figure~\ref{fig:pipeline}, we first collect a large number of real-world editing cases, balancing across MS-COCO object categories to obtain the raw dataset, denoted as $\mathcal{D}$. 
The key step in this process is dataset filtering. 
We observe that pretrained editing models, such as Qwen-Image-Edit-2509, already demonstrate strong editing capabilities in most scenarios. 
As illustrated by the reward curve in Figure~\ref{fig:reward}, the raw dataset $\mathcal{D}$ provides limited improvement during training because a substantial portion of cases receive the maximum score of~5, preventing the model from focusing on low-score, hard cases within a training batch. 
To increase the difficulty of RL post-training and better emphasize challenging cases, we discard the high-score samples (score~5) in $\mathcal{D}$ to construct the final dataset, denoted as $\mathcal{D}^\dagger$.

\textbf{Task-aware RL Post-Training.} 
In the third stage, based on $\mathcal{D}^\dagger$, the framework applies the HP-Scorer as the reward model and employs online Flow-GRPO for post-training. 
Specifically, the final reward is normalized to the range $[0,1]$ using a sigmoid function:
\begin{equation}
r = \frac{1}{1 + \exp(-\alpha*s + \beta)}
\end{equation}
\begin{equation}
s = \textit{HP-Scorer}(A, B, T, scoring ~ prompt)
\end{equation}
where the \textit{scoring prompt} is carefully designed for each task, as described in Stage~1 of the framework. 
Here, $\alpha$ and $\beta$ are scaling and shift parameters, set to $2$ and $5$, respectively. 

Through these three stages, HP-Edit efficiently post-trains the pretrained editing model to generate results that better align with high human-preference scores.

\subsection{Details of Dataset}
\begin{figure}
  % \begin{subfigure}{0.28\linewidth}
  \begin{subfigure}{1.0\linewidth}
    \includegraphics[width=1\linewidth]{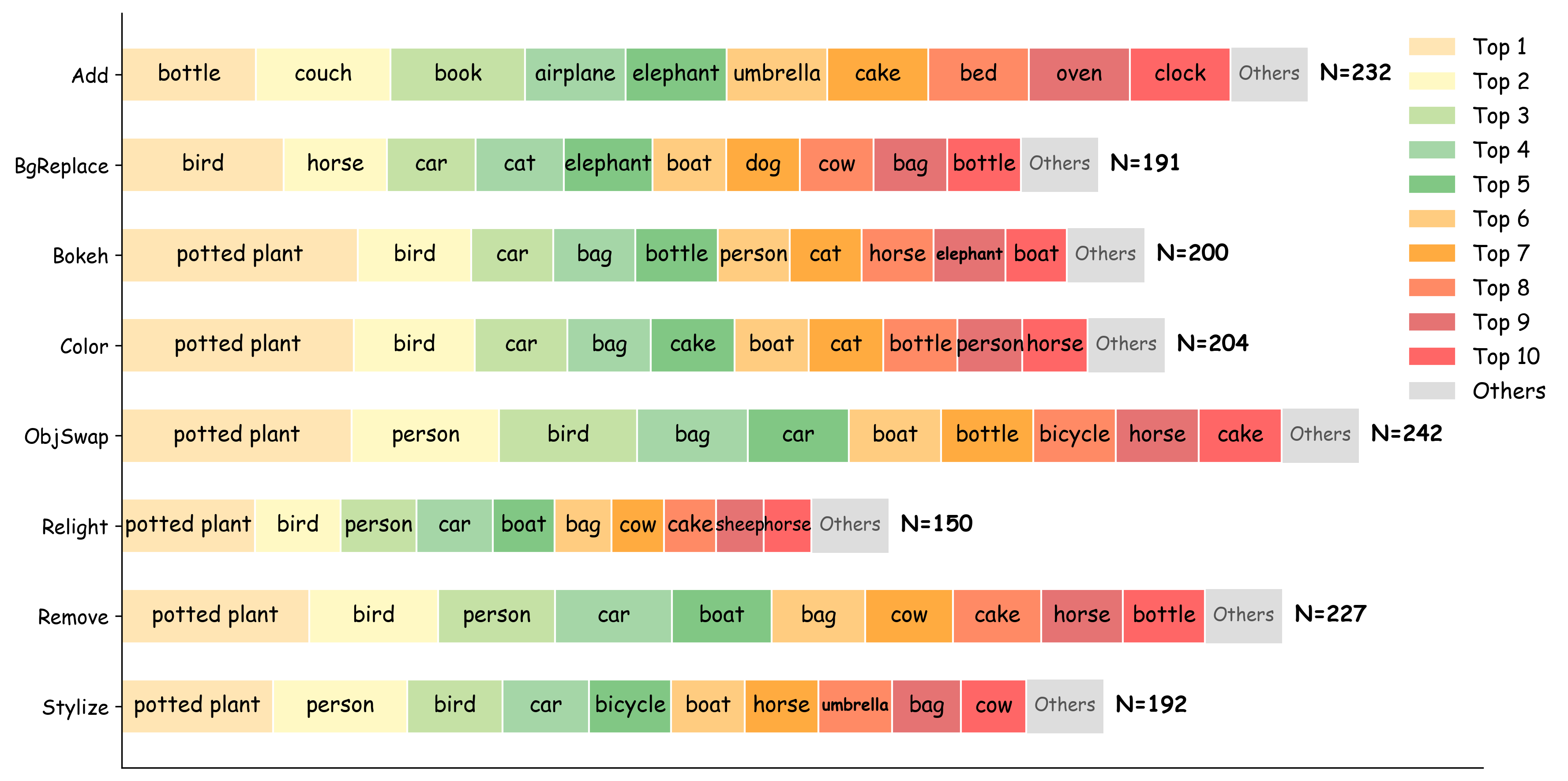}
    % \fbox{\rule{0pt}{2in} \rule{.9\linewidth}{0pt}}
    % \caption{Another example of a subfigure.}
  \end{subfigure}
  \caption{The details of task and object distribution in RealPref-50K}
  \label{fig:data}
\end{figure}

\textbf{RealPref-50K} focuses on real-world image scenarios and contains $55{,}795$ editing cases across eight common editing tasks, as shown in Figure~\ref{fig:data}: object addition, object removal, object swapping, background replacement, color change, bokeh (background blur), relighting, and style transfer. 
All source images are collected from high-quality, open-source real-world datasets (e.g., Pixabay~\cite{pixabay}, LSDIR~\cite{li2023lsdir}, and DIV2K~\cite{agustsson2017ntire}, etc.) to ensure visual diversity and realism. For editing instruction annotation, we employ a visual language model (VLM) to automatically generate textual editing instruction based on the input image. 
Subsequently, we calculate the CLIP score across all MS-COCO~\cite{lin2014microsoft} categories (e.g., ``\textit{person}'', ``\textit{car}'', ``\textit{bear}'', ``\textit{cake}'') to measure the similarity between input image and category embeddings. 
As illustrated in Figure~\ref{fig:data}, this process facilitates category balancing, ensuring a relatively uniform distribution of object categories across the dataset. 

Finally, image editing models (e.g., Qwen-Image-Edit-2509) are used to generate the corresponding edited outputs, forming triplets $(A, B, T)$, where $A$ is the input image, $B$ is the edited image, and $T$ is the instruction. 
All triplets are subsequently scored by the HP-Scorer to filter high-quality cases. 
As a result, \textbf{RealPref-50K} achieves balanced coverage across both common editing tasks and object categories, while focusing on hard editing cases that align with human rating in real-world scenarios

\subsection{Benchmark}
\textbf{RealPref-Bench} is a benchmark designed to evaluate image editing models using real-world images and manually verified editing instructions aligned with human preferences.
It contains $1{,}638$ cases, with approximately $200$ instances per sub-task. 
Both RealPref-Bench and RealPref-50K are curated to ensure balanced coverage of common MS-COCO object categories, ensuring consistency and representativeness across the evaluation and training domains.

% we apply task-aware GRPO to train editing models with the proposed human preference auto-scorer. Extensive experiments demonstrate that our HP-Edit framework efficiently improves editing models like FLUX.1-Kontext and Qwen-Image-Edit, enhancing image editing capabilities to align with specific human preferences in real-world scenarios.

% \subsubsection{RealPref-bench}

% \subsection{Task-ware post-training}

\section{Experiments}

\begin{figure*}[!t]
    \centering
    \includegraphics[width=\textwidth]{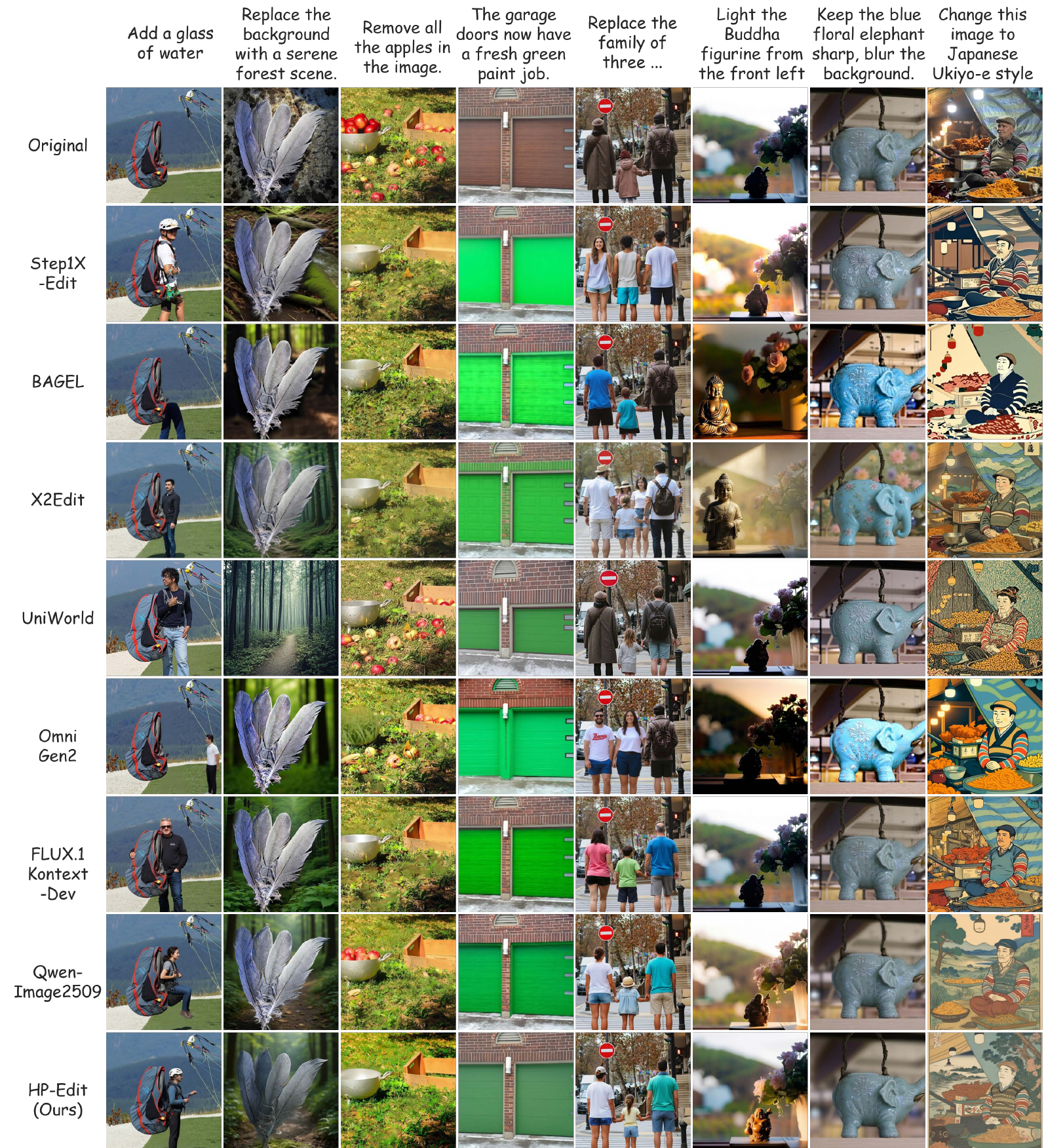} 
    \caption{Qualitative comparison of on the RealPref-Bench across eight common editing tasks.} 
    % \vspace{-0.3cm}
    \label{fig:main_results}
\end{figure*}

\begin{table*}[!th]
\caption{Quantitative comparison of HP-score produced on the proposed RealPref-Bench. Bold indicate the best performance.}
\resizebox{\linewidth}{!}{
\centering
% \footnotesize
\begin{tabular}{l|llllllll|l|l}
\toprule
% \hline
Model                & Add   & Remove & Obj. Swap & Bg. Replace & Color & Bokeh & Relighting & Style & Overall & Human \\ \hline
Step1X-Edit\cite{liu2025step1x}          & 4.612 & 3.934  & 4.174     & 4.696       & 3.922 & 3.445 & 3.407      & 4.37  & 4.07    &  3.89     \\
BAGEL\cite{deng2025Bagel}                & 3.547 & 3.727  & 3.74      & 4.257       & 4.162 & 3.335 & 2.78       & 3.651 & 3.65    &  3.588     \\
X2Edit\cite{ma2025x2edit}               & 4.19  & 2.841  & 3.657     & 3.686       & 3.338 & 1.695 & 3.107      & 4.443 & 3.396   &    3.355   \\
UniWorld-V1\cite{lin2025uniworld}          & 3.832 & 3.273  & 2.917     & 3.487       & 3.074 & 2.015 & 2.967      & 3.776 & 3.168   &    3.263   \\
Omnigen2\cite{wu2025omnigen2}            & 3.72  & 3.965  & 3.876     & 4.304       & 2.745 & 2.57  & 2.66       & 3.818 & 3.457   & 3.435      \\
Qwen-Image-Edit\cite{wu2025qwen}      & 4.151 & 3.665  & 4.574     & 4.539       & 4.549 & 3.02  & 1.98       & 4.875 & 3.919   &  4.005     \\
FLUX.1 Kontext-Dev\cite{labs2025flux}   & 4.431 & 4.388  & 4.116     & 4.23        & 3.99  & 2.28  & 1.247      & 4.047 & 3.59    &  3.345     \\ \hline
Qwen-Image-Edit-2509\cite{wu2025qwen} & 4.81  & 4.85   & 4.781     & 4.539       & 4.358 & 4.165 & 3.54       & 4.734 & 4.472   &   4.337    \\
HP-Edit  & \textbf{4.91}  & \textbf{4.925}  & \textbf{4.781}     & \textbf{4.733}       & \textbf{4.75}  & \textbf{4.545} & \textbf{3.913}      & \textbf{4.776} & \textbf{4.667}   &   \textbf{4.554}    \\ 
\bottomrule
\end{tabular}
}
\label{tab:main}
\end{table*}

\subsection{Experimental Settings}
\textbf{Experimental Setup.}
HP-Edit is a post-training framework for improving human-preference alignment. We use the open-source Qwen-Image-Edit-2509~\cite{wu2025qwen} as the base pretrained editing model to demonstrate the effectiveness of HP-Edit. Most parameters of the base model are frozen to preserve its pretrained capabilities, and we train only a lightweight LoRA with rank~$32$, using the AdamW optimizer~\cite{Loshchilov2019} with a learning rate of $3\times10^{-4}$. 
During training, the HP-Scorer employs the Qwen3-VL-32B-Instruct\footnote{Qwen/Qwen3-VL-32B-Instruct} instead of GPT-4o, as relying on external APIs leads to unstable latency and occasional failures, which can negatively affect the online RL training process.

\textbf{Evaluation.}
We first compare our method against state-of-the-art editing models on RealPref-Bench using the score produced by the HP-Scorer, referred to as  \textbf{HP-Score} . 
The comparison includes Step1X-Edit~\cite{liu2025step1x}, BAGEL~\cite{deng2025Bagel}, X2Edit~\cite{ma2025x2edit}, UniWorld-V1~\cite{lin2025uniworld}, OmniGen2~\cite{wu2025omnigen2}, Qwen-Image-Edit~\cite{labs2025flux}, FLUX.1-Kontext(Dev), and Qwen-Image-Edit-2509. 
For fair comparison across all methods, HP-Score is computed via GPT-4o based on HP-Scorer, which evaluates each edited result on a $0$–$5$ scale.

\subsection{Qualitative and Quantitative Results}
% We comprehensively evaluate our approach from two perspectives: the alignment between the proposed HP-Scorer scoring and human rating, and the performance of the editing model after applying HP-Edit. For quantitative evaluation, we use the proposed RealPref-Bench and GEdit-Bench~\cite{liu2025step1x}, which assess general-purpose image editing under diverse natural-language instructions. As shown in Figure~\ref{fig:main_results}, while FLUX.1-Kontext-dev and Qwen-Image-Edit-2509 exhibit excellent visual results in most cases, they still struggle to accurately follow user instructions, especially when editing large regions. In addition, BAGEL, UniWorld-V1, and Step1X-Edit sometimes produce obvious artifacts and distortions, lacking realism and fidelity. Qwen-Image-Edit also fails to consistently preserve key visual characteristics of subject identity. Overall, HP-Edit—both the FLUX and Qwen variants—achieves the best performance in terms of visual quality and instruction following.

% Additionally, we conduct comparisons on the open-source GEdit-Bench~\cite{liu2025step1x} using \textbf{G\_SC}, \textbf{G\_PQ}, and \textbf{G\_O} to demonstrate the effectiveness of our framework. As shown in Table~\ref{tab:GEdit}, \textbf{HP-Edit} not only delivers strong performance across all three metrics but also exhibits task-level generalization: although the tasks used to train HP-Edit are not identical to those in GEdit-Bench, it maintains—or improves—overall performance without degradation.
To evaluate the effectiveness of our method, we conduct experiments from two perspectives: (1) the alignment between different VLM-based scoring methods and human judgments, and (2) the performance improvement of the editing model after post-training with our method. 

\textbf{Qualitative Analysis.} 
As shown in Figure~\ref{fig:main_results}, HP-Edit produces results that are more faithful to the editing instructions while also exhibiting higher realism, fewer artifacts, and better preservation of scene structure. 
In contrast, baseline methods such as Step1X-Edit and UniWorld-V1 often introduce noticeable distortions under challenging edits (e.g., large-area removal or background replacement), and FLUX.1-Kontext-Dev occasionally generates unrealistic results with a painted or stylized appearance. 
The Qwen-Image-Edit-2509 baseline, while strong, still falls short on tasks requiring human aesthetic judgment—precisely the gap HP-Edit is designed to address.

\textbf{Quantitative Analysis}. Table~\ref{tab:main} reports the quantitative results on the proposed RealPref-Bench.
Overall, HP-Edit achieves the best performance across almost all eight editing sub-tasks as well as the overall HP-Score, outperforming both foundational editing models (e.g., Qwen-Image-Edit-2509 and FLUX.1-Kontext-Dev) and previous state-of-the-art methods such as Step1X-Edit, BAGEL, and X2Edit. 
Notably, HP-Edit achieves an overall score of \textbf{4.667}, improving upon the strong Qwen-Image-Edit-2509 baseline (4.472) by a significant margin. This clearly demonstrates the effectiveness of our human-preference post-training strategy in enhancing both instruction-following ability and visual quality.
To further validate generalizability, we evaluate HP-Edit on Geditbench—the official benchmark of Step1X-Edit \cite{liu2025step1x}, where it also achieves state-of-the-art performance, outperforming Step1X-Edit and other comparative methods. This confirms our human-preference alignment strategy transfers effectively to existing standard benchmarks.
From a task-level perspective, HP-Edit consistently ranks first across all eight categories.
The most pronounced improvements appear in tasks that require fine-grained appearance consistency or strong realism priors, such as \emph{color change}, \emph{bokeh}, \emph{relighting}, and \emph{background replacement}. These tasks typically involve subtle semantic reasoning or complex visual adjustments, where the pretrained models often struggle—yet HP-Edit shows clear gains, indicating that the preference-aligned reward effectively guides the model toward more human-desired outputs. \\
Together, these qualitative and quantitative results validate the effectiveness of HP-Edit in delivering human-preference–aligned editing improvements across diverse real-world scenarios.

\subsection{Ablation Study}
% As shown in Figure~\ref{fig:reward}, the blue curve (BaseData + BaseScorer) exhibits the highest initial reward because BaseData contains many easy, high-score cases, causing the model to receive saturated rewards without meaningful gradient feedback. In contrast, the yellow curve (RealPref-50K + BaseScorer) shows a clear improvement at the beginning of training. This is because RealPref-50K is filtered by the HP-Scorer to remove high-score trivial cases while retaining more low-score, hard examples, enabling the model to receive informative reward signals and focus on challenging edits.

% The green curve (RealPref-50K + HP-Scorer), corresponding to HP-Edit, further demonstrates a stable and consistent upward trend, surpassing the other two settings. The task-specific HP-Scorer provides more accurate and fine-grained preference feedback, allowing the model to progressively align with human-preference patterns. These results confirm that both filtering the data using HP-Scorer and upgrading BaseScorer to HP-Scorer are crucial for improving the reward dynamics and overall RL post-training effectiveness.
We compare three training settings to analyze the effects of RealPref-50K and HP-Scorer:  
\begin{itemize}
\item  \textbf{BaseData + BaseScorer}, which uses the unfiltered dataset together with a simple primary scoring prompt across different tasks;
\item \textbf{RealPref-50K + BaseScorer}, which applies filtered RealPref-50K while retaining the Basescorer;
\item \textbf{RealPref-50K + HP-Scorer}, which corresponds to our full framework, HP-Edit, combining both the filtered dataset and the task-specific human preference scorer.
\end{itemize}

\begin{figure}[!th]
  % \begin{subfigure}{0.28\linewidth}
  % \begin{subfigure}{1.0\linewidth}
  %   \fbox{\rule{0pt}{2in} \rule{.9\linewidth}{0pt}}
  %   % \caption{Another example of a subfigure.}
  % \end{subfigure}
  \resizebox{\columnwidth}{!}{
  \includegraphics[width=\textwidth]{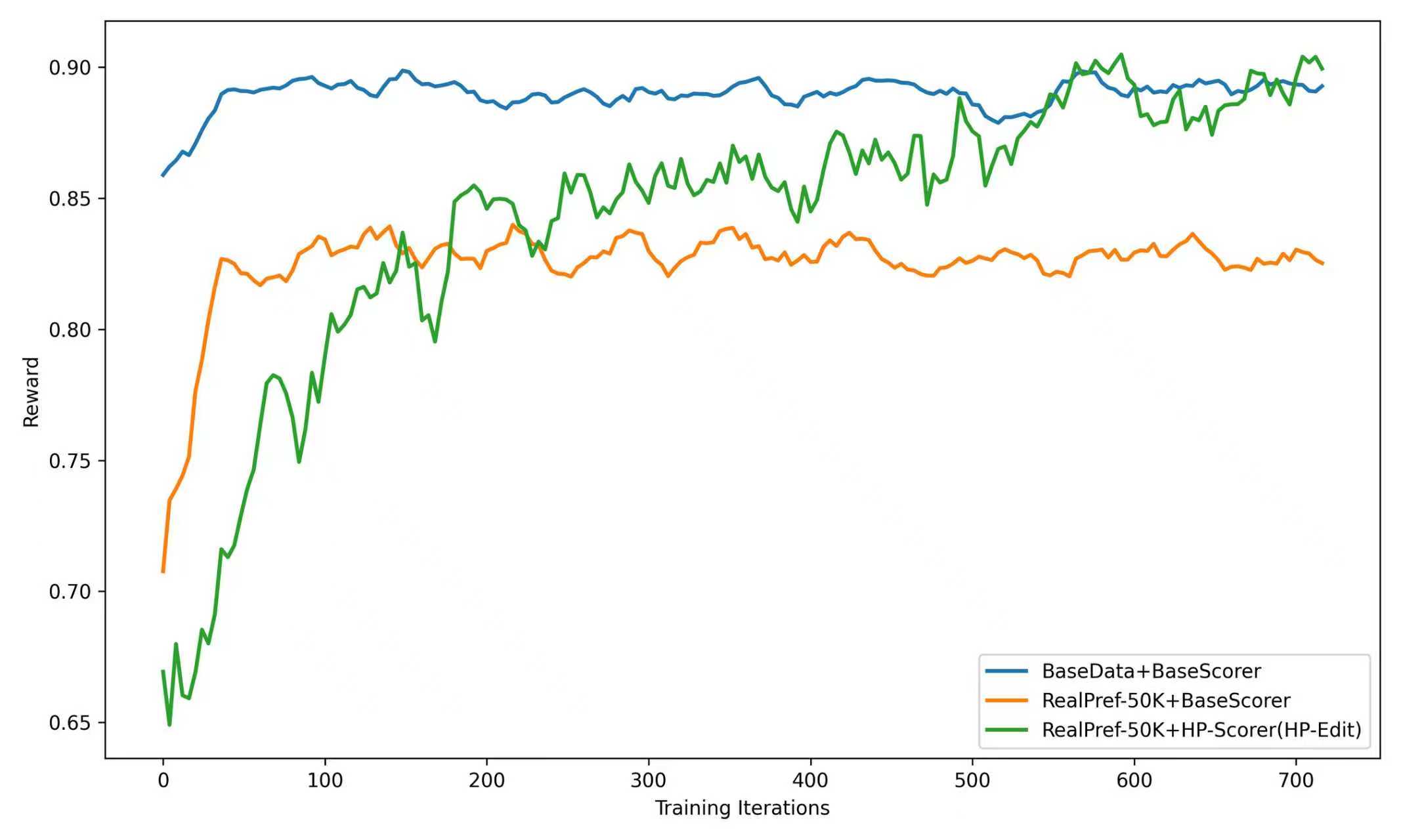}}
  \caption{Reward curves of HP-Edit with different settings.}
  \label{fig:reward}
\end{figure}

% \textbf{Reward Curve Analysis.}
As shown in Figure~\ref{fig:reward}, the blue curve (BaseData + BaseScorer) starts with the highest reward but shows minimal improvement, as BaseData contains many easy, high-score samples that lead to saturated rewards and weak training signals.
In contrast, the yellow curve (RealPref-50K + BaseScorer) demonstrates a clear upward trend at the beginning of training. 
By filtering out trivial high-score cases and retaining more low-score, hard examples, RealPref-50K provides more informative gradients and facilitates faster reward gains. The green curve (RealPref-50K + HP-Scorer), corresponding to HP-Edit, shows the most stable and consistent upward trajectory. 
% The task-specific HP-Scorer offers more fine-grained preference feedback, allowing the model to better capture human-preferred editing behaviors. 
These results validate that both dataset filtering and scorer refinement are essential for effective human-preference-aligned RL post-training.

% \textbf{Ablation for auto-scorer}. The human-preference auto-scorer, based on a pretrained VLM, is the core module of HP-Edit. It is used both to filter hard cases that align with human preferences and to serve as a task-aware reward function. As shown in Table~\ref{tab:framework}, we conduct an ablation study for HP-Edit based on Qwen-Image-Edit-2509. Specifically, \textbf{Base-Scorer} denotes a simple scoring prompt (\textit{``Please output the final evaluation score (0–5) for the editing task.''}) applied uniformly across tasks, whereas \textbf{Auto-Scorer} uses an optimized, task-specific preference-scoring prompt (details in the Supplementary). The model sizes \textbf{32B} and \textbf{235B} correspond to \textit{Qwen3-VL-32B-Instruct} and \textit{Qwen3-VL-235B-A22B-Instruct}, respectively. The results show that Auto-Scorer consistently outperforms Base-Scorer under the same VLM, and that performance improves as the VLM size increases.
% \textbf{Quantitative Analysis.}
Table~\ref{tab:ablation} further quantifies the contributions of RealPref-50K and the HP-Scorer. 
Using the unfiltered BaseData with the simple BaseScorer results in a slight performance drop compared to the pretrained baseline (4.391 vs.\ 4.472), indicating that the raw dataset contains many overly easy or noisy cases that provide weak or misleading RL signals. 
Replacing BaseData with the filtered RealPref-50K yields a noticeable improvement (4.577), demonstrating that removing trivial high-score samples and emphasizing harder, low-score cases enables more effective preference learning. 
Finally, combining RealPref-50K with the task-specific HP-Scorer (our full HP-Edit framework) achieves the highest score of \textbf{4.667}, outperforming all other settings. 
This consistent gain confirms that both components—high-quality preference-focused data and a task-aware preference scorer—are essential for maximizing alignment with human preferences during RL post-training.

As shown in Table~\ref{tab:GEdit}, our methods demonstrate superior performance on GEdit-Bench-EN~\cite{liu2025step1x}. We also provide a correlation analysis between the HP-score and user scores for GEdit-Bench-EN; further details are available in the supplementary materials.

\begin{table}[th]
\centering
\caption{
Ablation Study of the HP-Edit on the RealPref-Bench.}
\resizebox{\columnwidth}{!}{
\begin{tabular}{l|ll}
% \hline
\toprule
Framework                               & HP-score \\ \hline
Baseline &   4.472           \\
BaseData + Base-Scocer              &  4.391           \\
RealPref-50K + Base-Scorer          &  4.577            \\ \hline
RealPref-50K + AutoPref-Scorer  (HP-Edit)         &  \textbf{4.667}           \\
\bottomrule
\end{tabular}
}
\label{tab:ablation}
\end{table}

% \textbf{Ablation for dataset}.
% As shown in the first two rows of Table~\ref{tab:framework}, \textbf{RealPref-50K} further improves HP-Edit’s performance under the same scorer. In addition, Figure~\ref{fig:reward} shows that the reward curve for \textit{BaseData} oscillates severely, indicating that the auto-scorer fails to function effectively as a reward model. By contrast, when we filter out easy cases from \textit{BaseData} and construct \textbf{RealPref-50K}, the reward increases steadily under the same auto-scorer, demonstrating stable and meaningful learning.

% \textbf{Ablation for LoRA-Rank}
% As aforementioned, HP-Edit adopts LoRA with rank~$32$, whose effect is ablated in Figure~\ref{fig:LoRA}. From a comprehensive comparison, we observe that performance improves steadily as the rank increases from 8 to 32, but remains largely unchanged once the rank exceeds 32, for both Qwen-based and FLUX-based HP-Edit.

% \begin{figure}[th]
%   % \begin{subfigure}{0.28\linewidth}
%   \begin{subfigure}{1.0\linewidth}
%     \fbox{\rule{0pt}{2in} \rule{.9\linewidth}{0pt}}
%     % \caption{Another example of a subfigure.}
%   \end{subfigure}
%   \caption{Ablation study for different LoRA ranks on RealPref-Bench.}
%   \label{fig:LoRA}
% \end{figure}

\subsection{User Study}
% We recruited five participants to evaluate the editing outputs of HP-Edit and baseline methods on RealPref-Bench, covering over 1k editing pairs. The evaluation focuses on two aspects: instruction following and image quality. Scores range from 0 to 5, where 0 indicates a complete failure (e.g., the edited result is identical to the input), 3 indicates that the instruction is mostly followed but the generated content lacks aesthetic quality or fidelity, and 5 indicates full instruction compliance with high image quality. These scoring criteria are consistent with those used in Section 4.1. As shown in Figure~\ref{fig:user}, we report the average score of each model on the benchmark. The results demonstrate that our method aligns strongly with human preferences in real-world scenarios.

\begin{figure}[th]
  % \begin{subfigure}{0.28\linewidth}
  % \begin{subfigure}{1.0\linewidth}
  %   \fbox{\rule{0pt}{2in} \rule{.9\linewidth}{0pt}}
  %   % \caption{Another example of a subfigure.}
  % \end{subfigure}
  \resizebox{\columnwidth}{!}{
  \includegraphics[width=\textwidth]{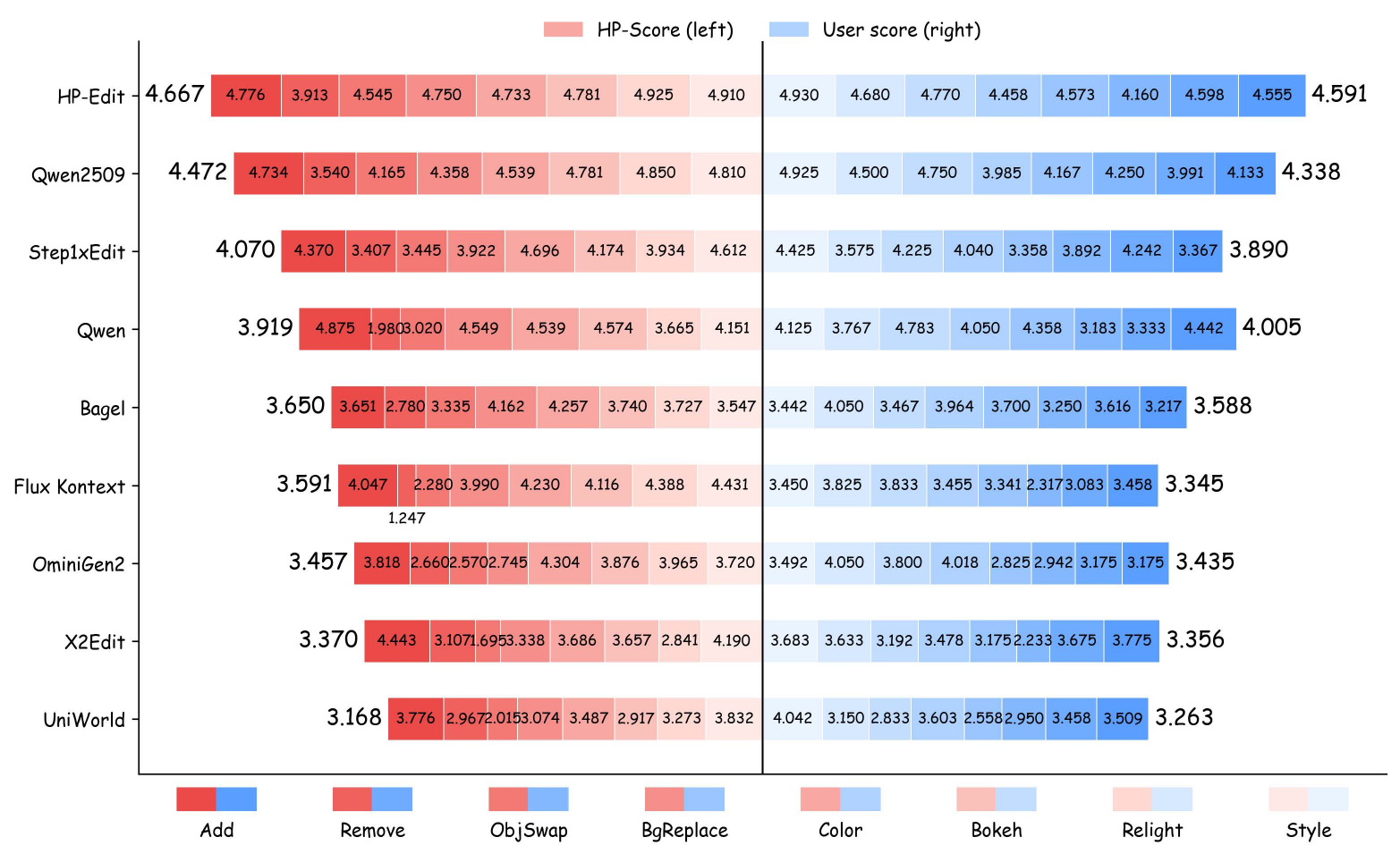}}
  \caption{HP-score and user score}
  \label{fig:user}
\end{figure}

We recruited five annotators to evaluate the editing outputs of HP-Edit and baseline methods on RealPref-Bench, covering over 1k editing pairs. The evaluation focuses on two aspects: instruction adherence and image quality. Scores range from 0 to 5, where 0 indicates a complete failure (e.g., the edited result is identical to the input), 3 indicates that the instruction is mostly followed but the generated content lacks aesthetic appeal or realism, and 5 indicates full instruction compliance with high image quality. These scoring criteria are consistent with those outlined in Section 4.1. As shown in Figure~\ref{fig:user}, we report the average score of each model on the benchmark. The score distribution from the user study closely matches the results produced by HP-Scorer. Across all tasks, we observe consistent improvements over the pretrained model, which not only demonstrates the effectiveness of HP-Edit but also validates the scoring accuracy and human alignment of HP-Scorer.
% For more details, please refer to Appendix~B.

% \usepackage{multirow}
\begin{table}[th]
\caption{Performance of different methods on GEdit-Bench-EN.}
\resizebox{\columnwidth}{!}{
\footnotesize
\begin{tabular}{l|lll}
\toprule
Model                      & \multicolumn{3}{l}{GEdit-Bench-EN ↑} \\ \cline{2-4} 
                                            & G\_SC       & G\_PQ      & G\_O      \\ \hline
Step1X-Edit\cite{liu2025step1x}          & 7.66        & 7.35       & 6.97      \\
BAGEL\cite{deng2025Bagel}                & 7.36        & 6.83       & 6.52      \\
X2Edit\cite{ma2025x2edit}               &6.80             & 8.37           & 7.03          \\
UniWorld-V1\cite{lin2025uniworld}          & 4.93        & 7.43       & 4.85      \\
Omnigen2\cite{wu2025omnigen2}            & 7.16        & 6.77       & 6.41      \\
Qwen-Image-Edit\cite{wu2025qwen}      & 8.00        & 7.86       & 7.56      \\ 
FLUX.1-Kontext-Dev\cite{labs2025flux}   & 6.52        & 7.38       & 6.00      \\ \hline
Qwen-Image-Edit-2509\cite{wu2025qwen} & 8.15        & 7.86       & 7.54      \\
HP-Edit              & \textbf{8.35}       &\textbf{8.54}     & \textbf{8.30}      \\ \bottomrule
\end{tabular}
}
\label{tab:GEdit}
\end{table}

% Research on image editing centers on controllability, local fidelity, and tight coupling with strong base generators. Data-driven fine-tuning and scalable pipelines\cite{brooks2023instructpix2pix,zhang2023magicbrush, li2024brushedit} improve editing realism, while architectural and adaptation techniques \cite{huang2024smartedit,shi2024seededit,yu2025anyedit,zhao2024ultraedit} enhance fine-grained control and efficiency. Unified frameworks \cite{xiao2025omnigen,wu2025omnigen2,han2024ace,mao2025ace++,liu2025step1x,deng2025Bagel,ma2025x2edit,lin2025uniworld} merge generation and editing into a unified stack. FLUX.1-Kontext \cite{labs2025flux} adds high-consistency reference conditioning for rapid, reference-guided edits. Qwen-Image \cite{wu2025qwen} leverages large-scale curriculum training for precise edits and complex text rendering.

\section{Conclusion and Limitation}
In this paper, we propose HP-Edit, a post-training framework for human-preference-aligned editing, and introduce RealPref-50K, a large-scale ``human-preference'' dataset, alongside the RealPref-Bench benchmark. Notably, RealPref-50K utilizes HP-Scorer as a scalable proxy for hard-case filtering and comprises high-quality pseudo-labels generated by the scorer. Despite its strengths, HP-Edit still struggles with code-switching or mixed Chinese-English text editing (e.g., `Translate the English text into Chinese'), a limitation largely inherited from the base models. We aim to address these challenges in future research.

%(e.g., `Translate the English text into Chinese')

% We admit that ``human-preference'' mainly refers to the alignment criteria derived from humans via HP-Scorer rather than manual labeling for the entire dataset. As noted by the reviewer, a core contribution of HP-Edit is leveraging HP-Scorer as a scalable proxy for hard-case filtering, which is essential for efficient RL post-training. In the final version, we will explicitly clarify that RealPref-50K largely consists of pseudo-labels generated by HP-Scorer. Upon the release of the dataset, we will clearly flag human-annotated vs. pseudo-labeled samples to ensure transparency.

{
    \small
    \bibliographystyle{ieeenat_fullname}
    \bibliography{main}
}


% \documentclass[10pt,twocolumn,letterpaper]{article}
% \usepackage{orcidlink}
\clearpage
\setcounter{page}{1}
\maketitlesupplementary

% Support for ORCID icon

\appendix % 标记补充材料开始
\renewcommand{\thesection}{S\arabic{section}} % 在章节编号
\setcounter{section}{0} % 重置章节计数器
\setcounter{figure}{0}  % 重置图计数器
\setcounter{table}{0}   % 重置表计数器

% \usepackage{orcidlink}
% \usepackage{enumitem}
% \newcounter{suppfigure}
\renewcommand{\thefigure}{S\arabic{figure}} 
% \newcommand{\suppfigref}[1]{S\ref{#1}}

% \DeclareCaptionLabelFormat{supp}{#1 S#2}
% \captionsetup[figure]{labelformat=supp}

\renewcommand{\thetable}{S\arabic{table}} 
% \newcommand{\supptabref}[1]{S\ref{#1}}

% \DeclareCaptionLabelFormat{supp}{#1 S#2}
% \captionsetup[table]{labelformat=supp}

% \usepackage{orcidlink}
% \usepackage{enumitem}
% \newcounter{suppfigure}
% \renewcommand{\thesuppfigure}{S\arabic{suppfigure}} 
% \newcommand{\suppfigref}[1]{S\ref{#1}}
% \newcounter{supptable}
% \renewcommand{\thesupptable}{S\arabic{supptable}}
% \newcommand{\supptabref}[1]{S\ref{#1}}

% \DeclareCaptionLabelFormat{supp}{#1 S#2}
% \captionsetup[figure]{labelformat=supp}
% \captionsetup[table]{labelformat=supp}

% \newcounter{mysec}[section]
% \renewcommand{\themysec}{\Roman{mysec}}
% \newcommand{\mysec}[1]{
%   \refstepcounter{mysec}
%   \subsection*{\themySection#1}
% }
% \maketitle
% % \input{sec/7_suppl}
% {
%     \small
%     \bibliographystyle{ieeenat_fullname}
%     % \bibliography{supplementary_material}
% }
% \begin{document}

Section~\ref{sec:detailmethod} provides more details of experiments of the main paper. Section~\ref{sec:detailprompt} supplements the details of the system prompts of HP-scorer per task.  Section~\ref{sec:quant} presents more quantitative comparisons. Section~\ref{sec:visual} presents more visual examples for qualitative comparisons. Section~\ref{sec:data} provides more details of RealPref-50k and RealPref-Bench.

\section{Experimental details}
\label{sec:detailmethod}
There are some annotation mistakes in the cases presented in Figure 4 of the main paper. The correct and complete instructions are shown below, where the order of cases 1–8 corresponds to Figure 4 from left to right.
\begin{itemize}
    \item case~1: ``\textit{Add a person standing on the green turf next to the paragliding harness, wearing a white helmet, holding the paraglider's control lines.}''
    \item case~2: ``\textit{Replace the background with a serene forest scene. The new background should have a path winding through tall trees, with lush green undergrowth. Ensure the lighting in the forest scene is a gentle glow to highlight the feathers.}
    \item case~3: ``\textit{Remove all apples in the image.}''
    \item case~4: ``\textit{Change the color of both garage doors from brown to green.}''
    \item case~5: ``\textit{Replace the family of three with a group of three people dressed in summer attire, such as shorts and t-shirts, while keeping the background and setting unchanged.}''
    \item case~6: ``\textit{Adjust the lighting so that the Buddha figurine is illuminated from the front-left, casting a soft shadow to the right and slightly behind, with a warm, diffused light source that enhances its surface details and creates gentle highlights on its rounded form.}
    \item case~7: ``\textit{Keep the the light blue ceramic elephant with floral patterns sharp, blur the background.}''
    \item case~8: ``\textit{Change this image to Japanese Ukiyo-e style, flat perspective, woodblock print texture, traditional Japanese colors, elegant composition, nature elements, cultural motifs, refined details, harmonious balance, ukiyo-e facial expressions, ukiyo-e landscape motifs.}''
\end{itemize}

\section{System prompts of HP-scorer for each task}
\label{sec:detailprompt}
HP-Scorer is highly dependent on the designed system prompts for evaluating the editing tasks, all of which are shown in Figure~\ref{fig:spl1_remove},\ref{fig:spl1_add},\ref{fig:spl1_swap},\ref{fig:spl1_bgreplace},\ref{fig:spl1_bokeh}, \ref{fig:spl1_relight},\ref{fig:spl1_style},\ref{fig:spl1_color}.

\begin{figure*}[!t]
    \centering
    \includegraphics[width=0.9\textwidth]{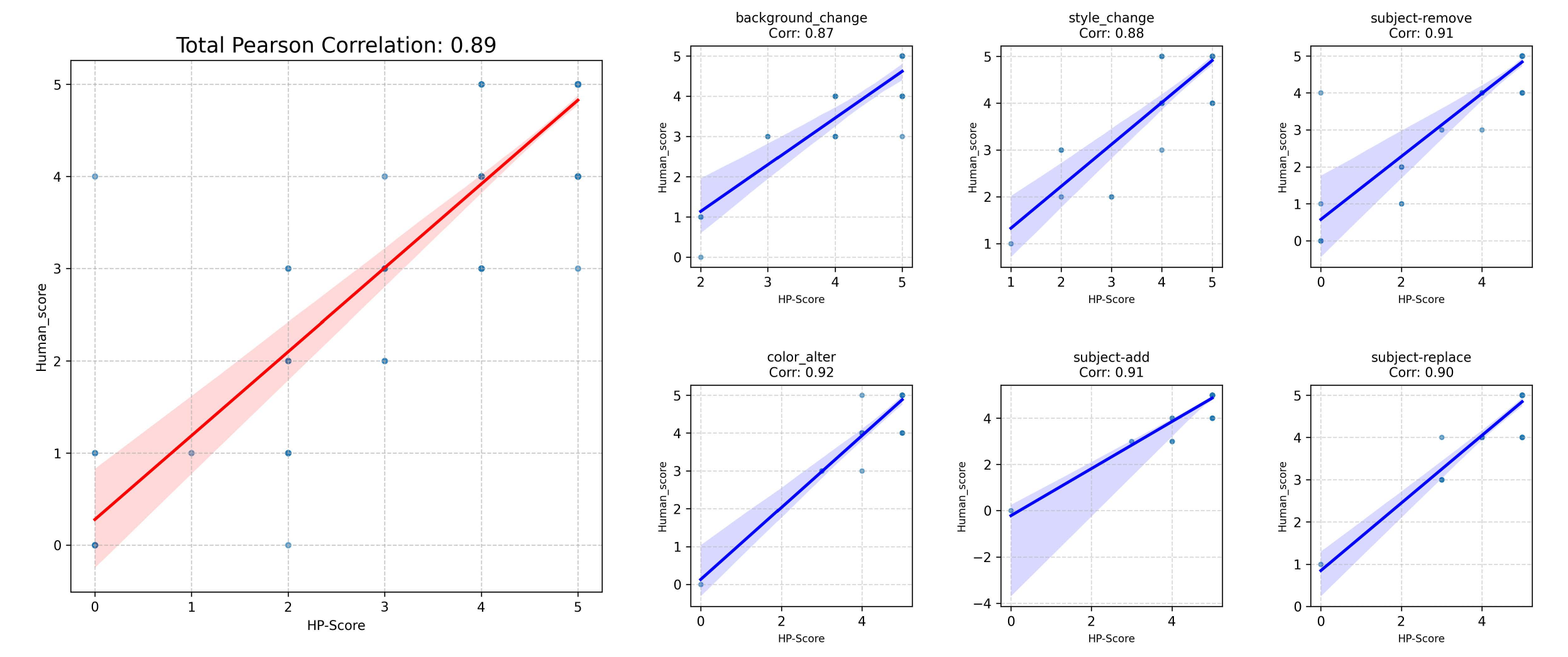}
  \caption{Correlation analysis between user score and HP-Score on GEdit-Bench-EN.}
  \label{fig:validity}
\end{figure*}

\begin{figure*}[!b]
    \centering
    \includegraphics[width=\textwidth]{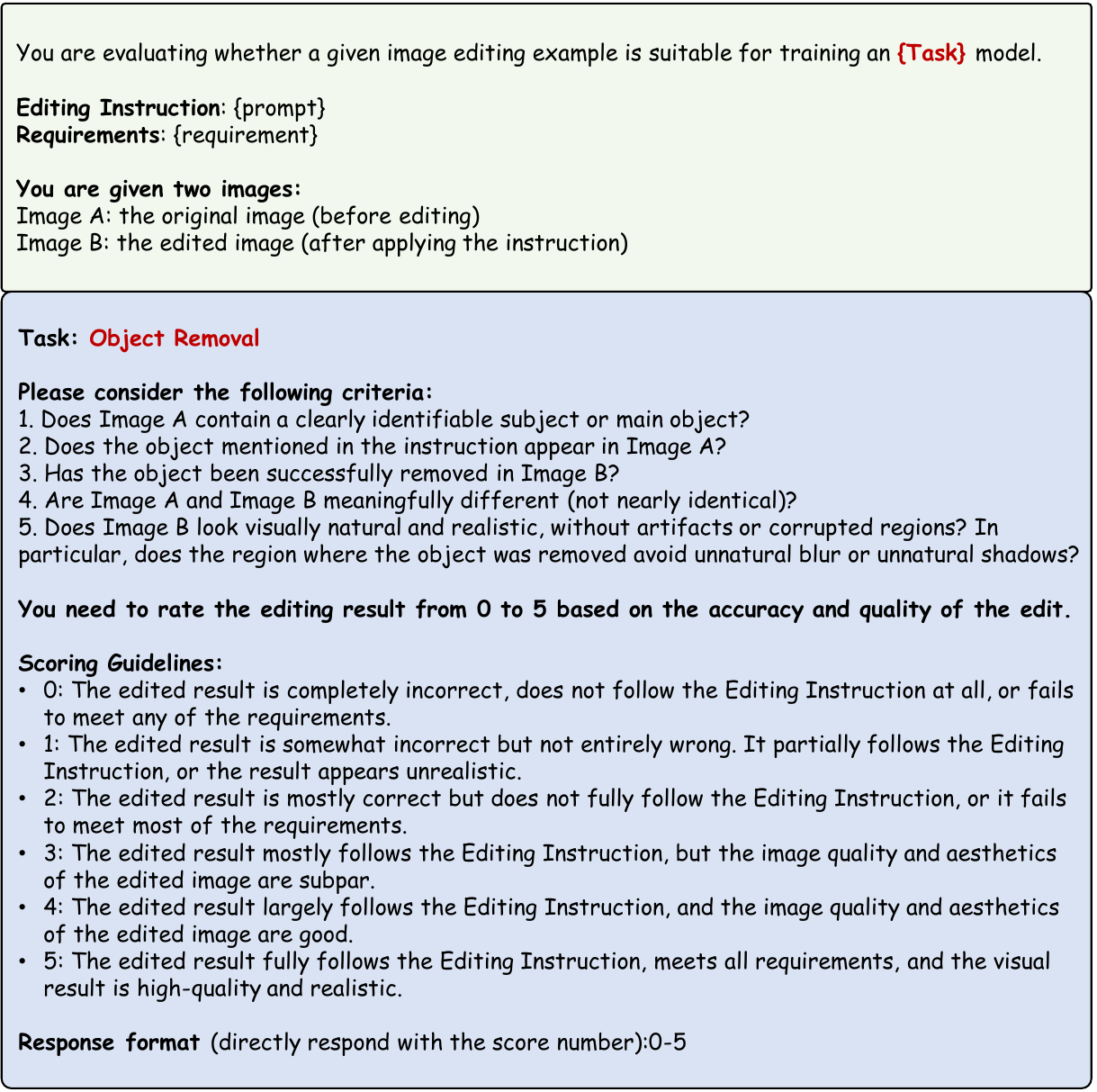} 
    \caption{System prompts of object removal task.} 
    \vspace{-0.3cm}
    \label{fig:spl1_remove}
\end{figure*}

\begin{figure*}[!b]
    \centering
    \includegraphics[width=\textwidth]{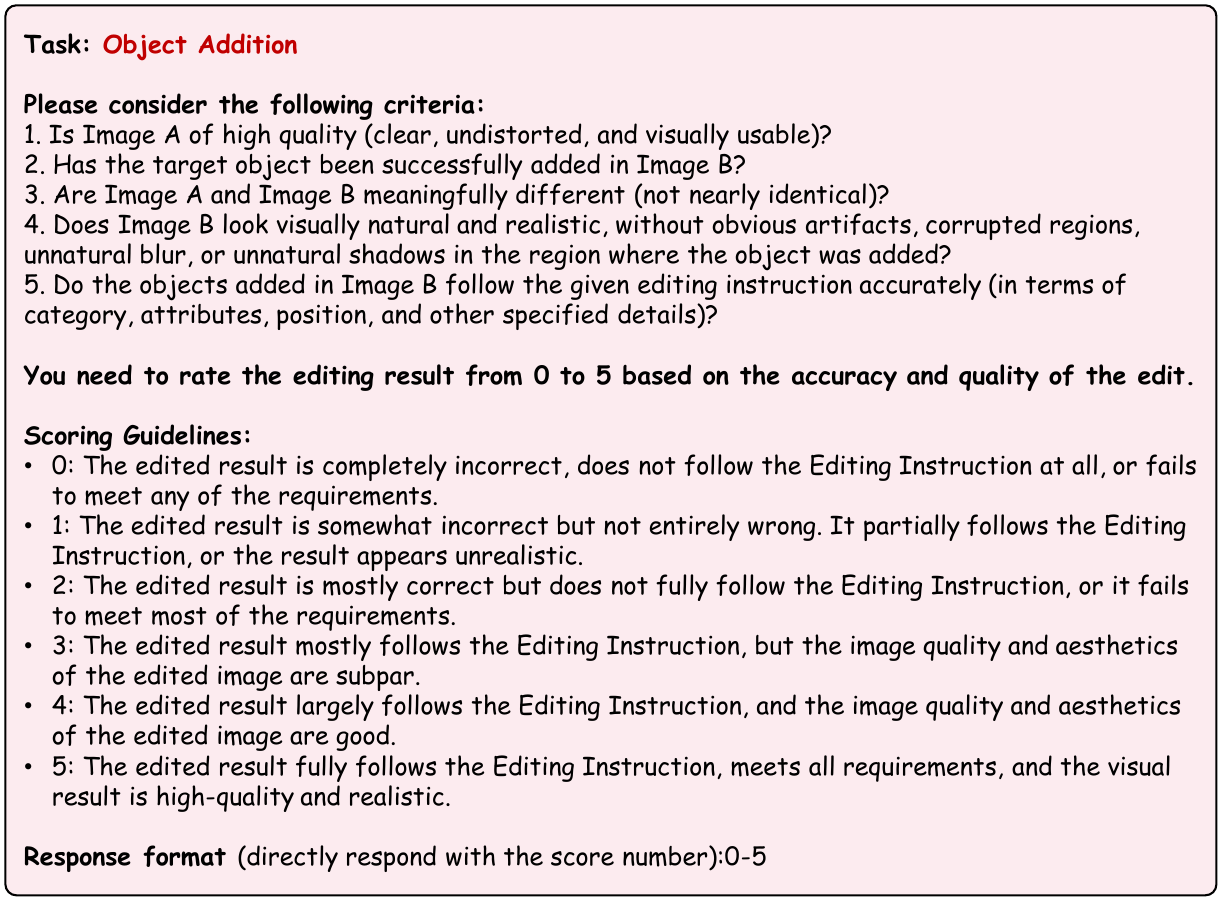} 
    \caption{System prompts of object adding task.} 
    \vspace{-0.3cm}
    \label{fig:spl1_add}
\end{figure*}

\begin{figure*}[!t]
    \centering
    \includegraphics[width=\textwidth]{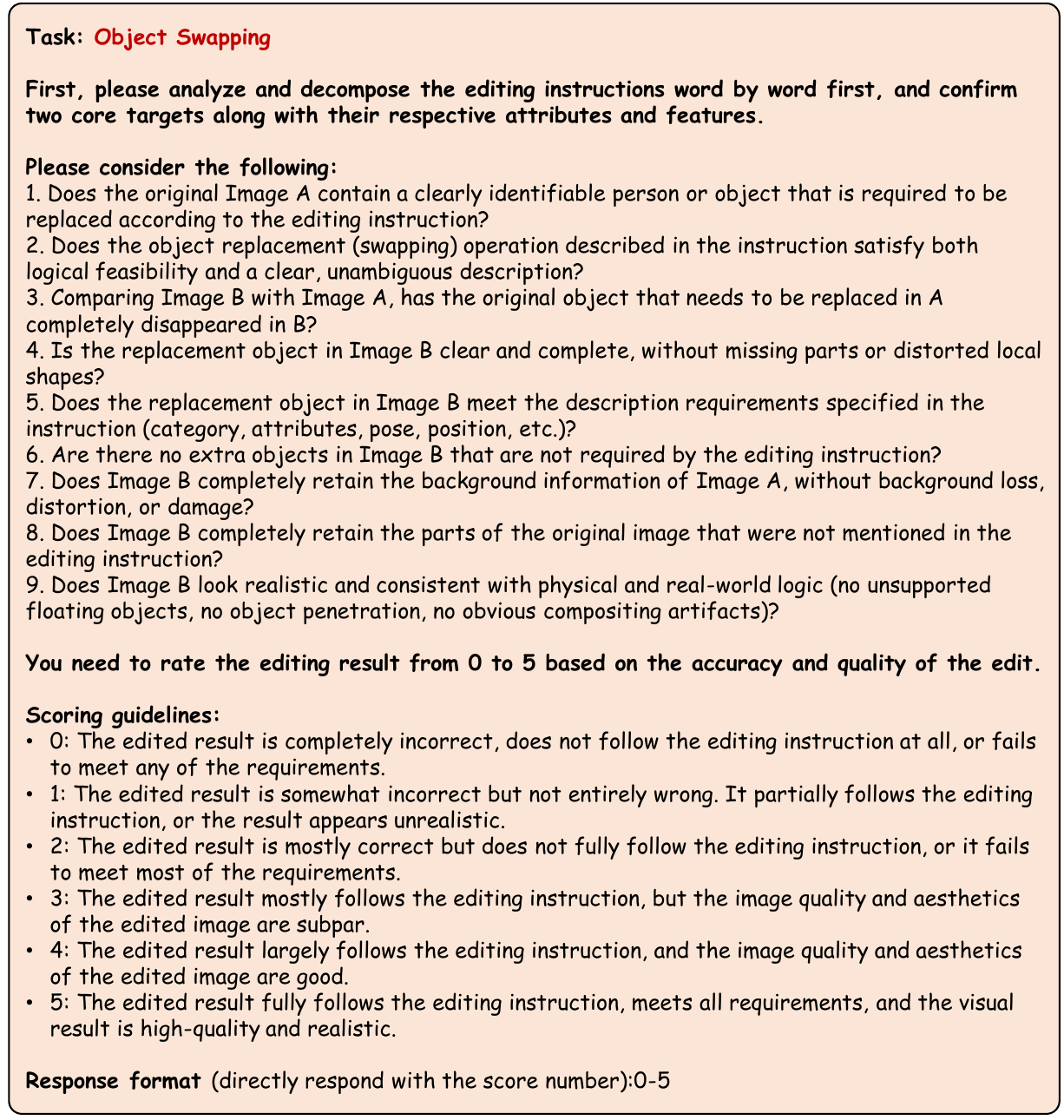} 
    \caption{System prompts of object swapping task.} 
    \vspace{-0.3cm}
    \label{fig:spl1_swap}
\end{figure*}

\begin{figure*}[!t]
    \centering
    \includegraphics[width=\textwidth]{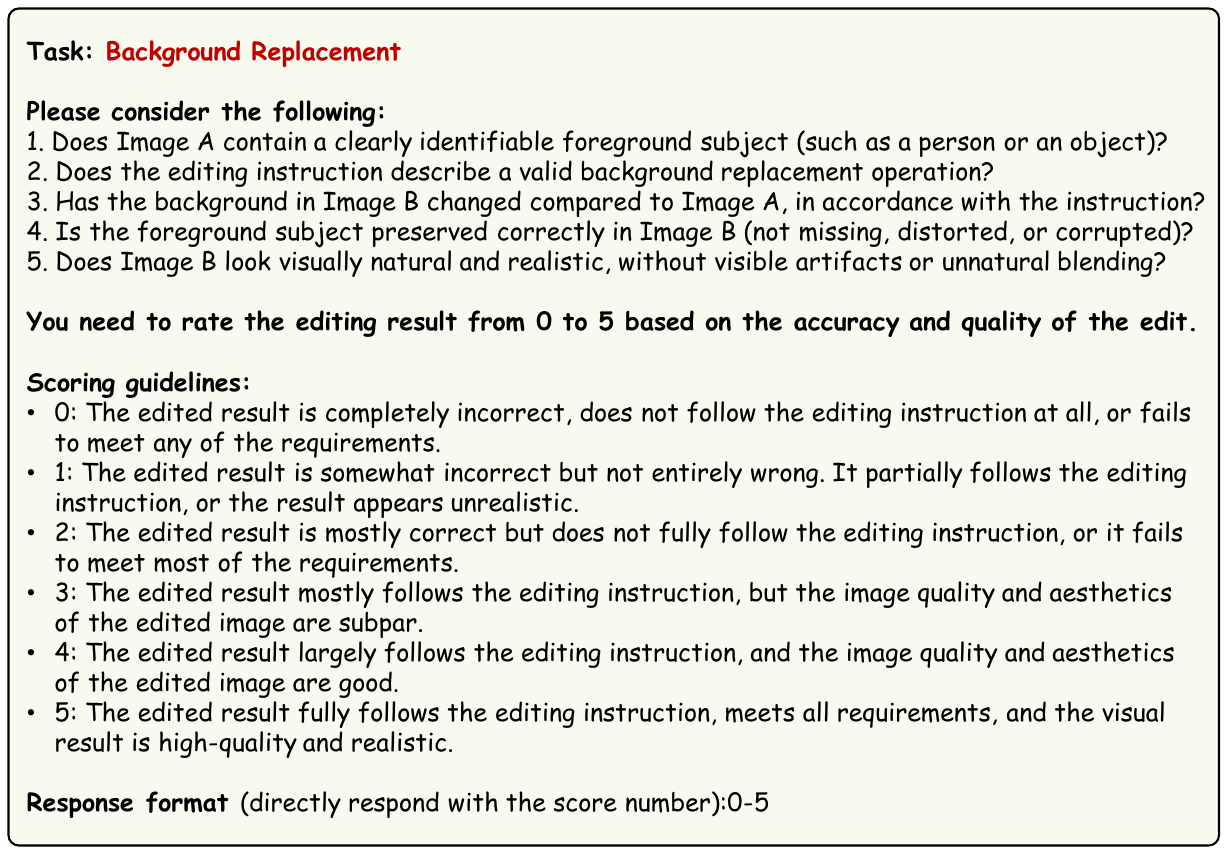} 
    \caption{System prompts of background replacement task.} 
    \vspace{-0.3cm}
    \label{fig:spl1_bgreplace}
\end{figure*}

\begin{figure*}[!t]
    \centering
    \includegraphics[width=\textwidth]{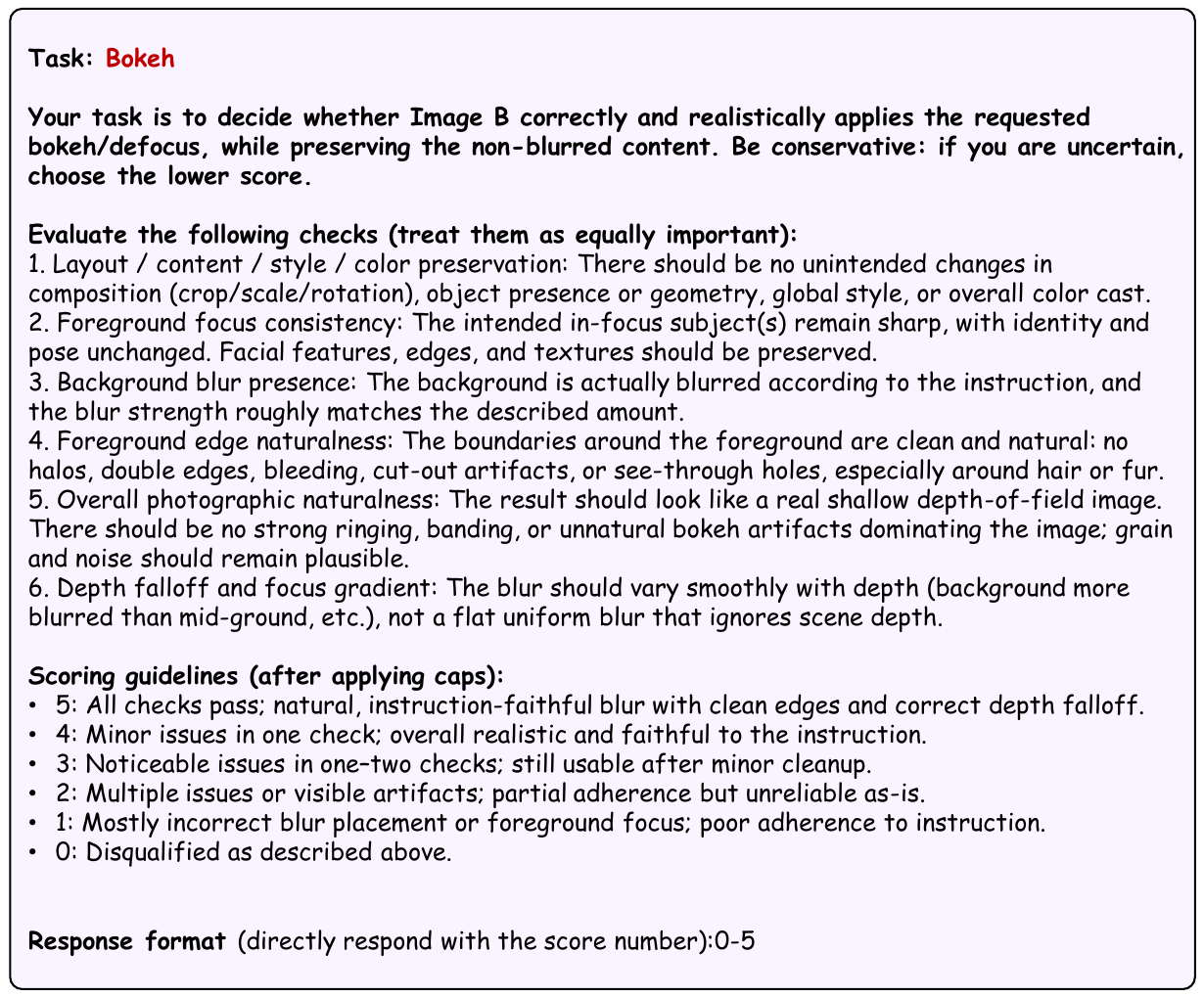} 
    \caption{System prompts of bokeh task.} 
    \vspace{-0.3cm}
    \label{fig:spl1_bokeh}
\end{figure*}

\begin{figure*}[!t]
    \centering
    \includegraphics[width=\textwidth]{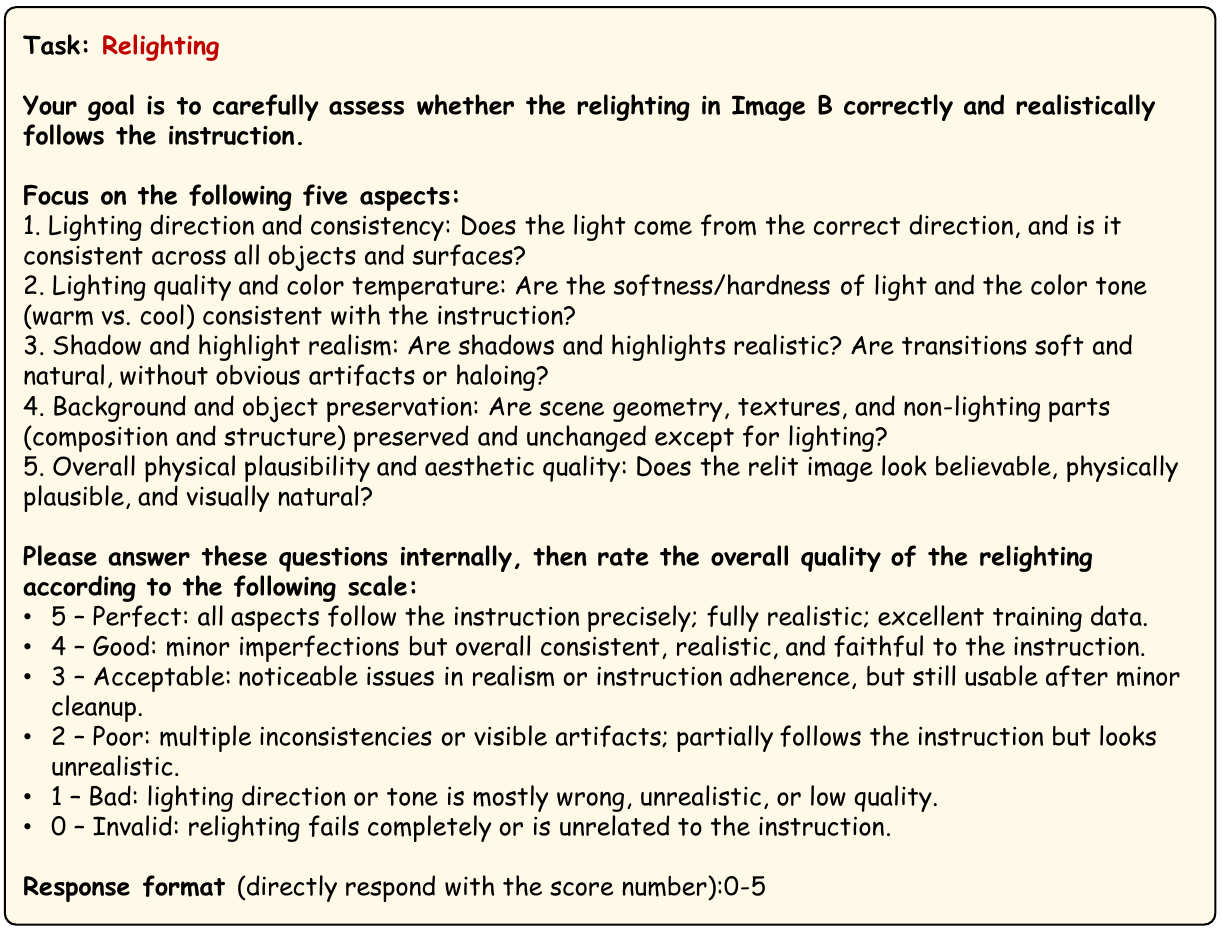} 
    \caption{System prompts of relighting task.} 
    \vspace{-0.3cm}
    \label{fig:spl1_relight}
\end{figure*}

\begin{figure*}[!t]
    \centering
    \includegraphics[width=\textwidth]{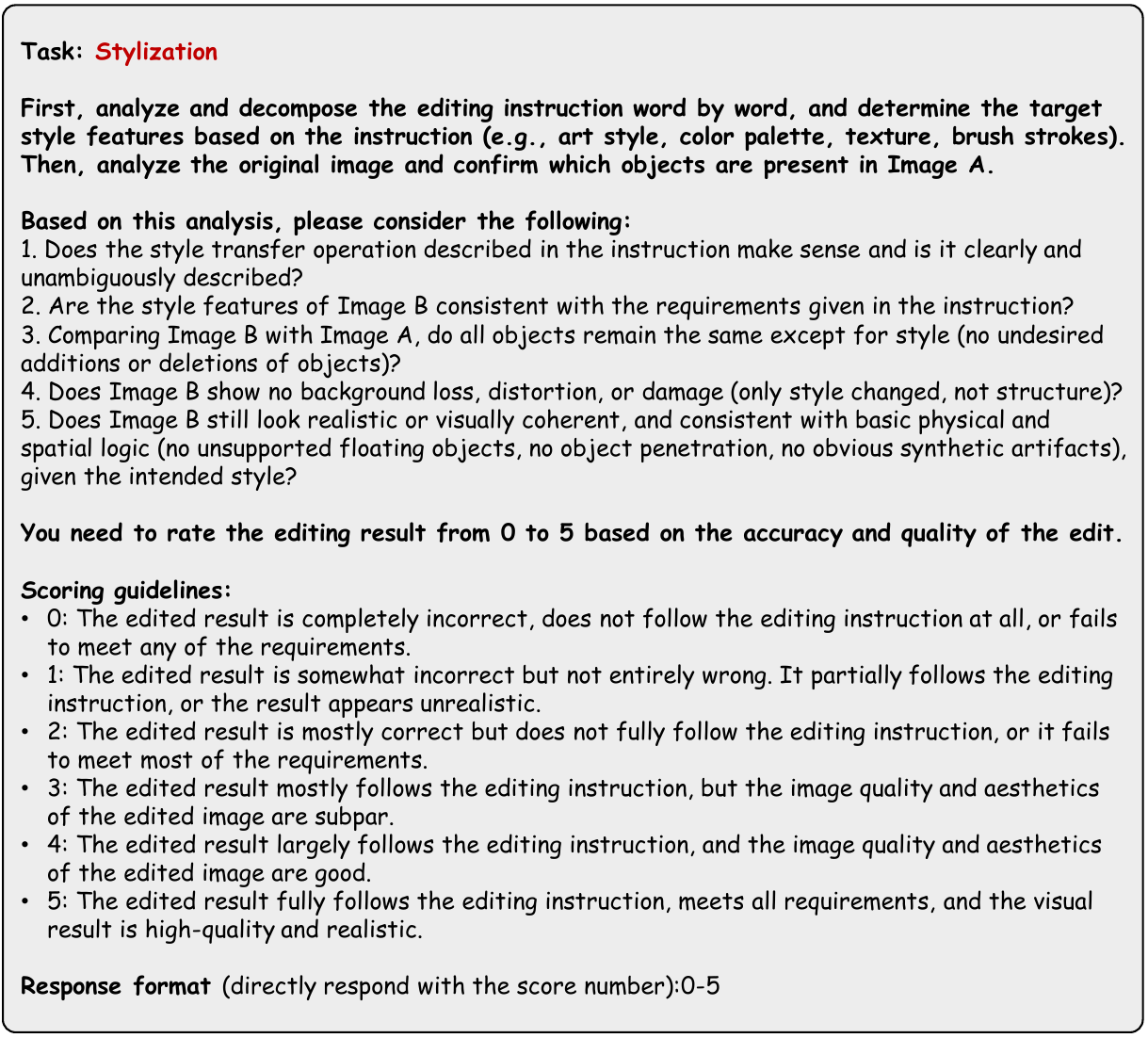} 
    \caption{System prompts of style changing task.} 
    \vspace{-0.3cm}
    \label{fig:spl1_style}
\end{figure*}

\begin{figure*}[!t]
    \centering
    \includegraphics[width=\textwidth]{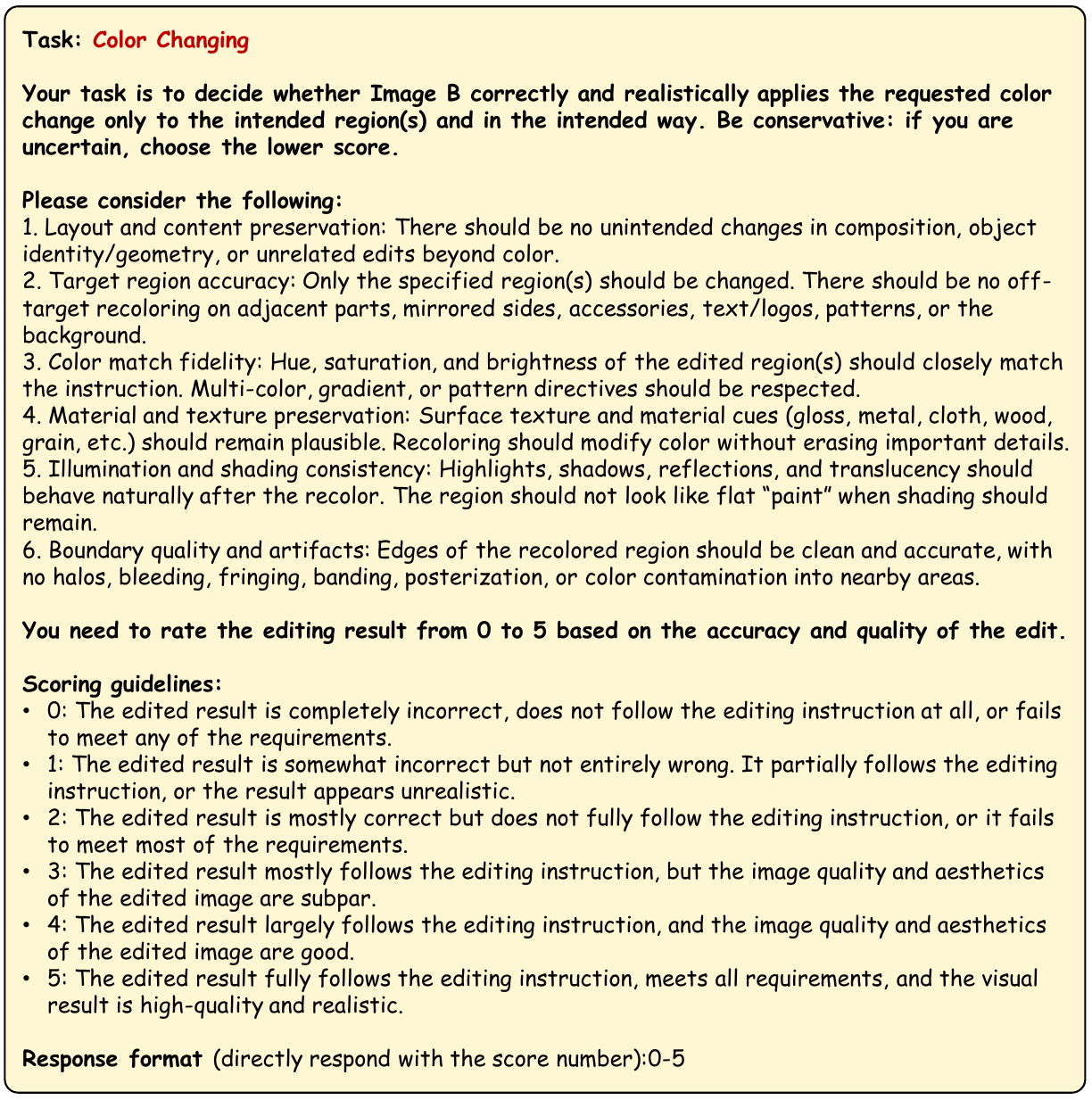} 
    \caption{System prompts of color changing task.} 
    \vspace{-0.3cm}
    \label{fig:spl1_color}
\end{figure*}

\section{More quantitative comparisons}
\label{sec:quant}
\textbf{Comparisons on different LoRA ranks}. As mentioned earlier, HP-Edit adopts LoRA with rank~$32$, and its effect is ablated in Figure~\ref{tab:LoRA}. We observe that performance improves steadily as the rank increases from 8 to $32$, but remains largely unchanged or begins to decline once the rank exceeds $32$.

\begin{table}[h]
\centering
\caption{
Comparison of HP-Edit with different LoRA ranks on the RealPref-Bench.
}
% \resizebox{\columnwidth}{!}{
\begin{tabular}{l|ll}
% \hline
\toprule
Methods                               & HP-score \\ \hline
HP-Edit (rank=8) &         4.614    \\
HP-Edit (rank=32)          &  4.667           \\
HP-Edit (rank=128)         &  4.645    \\
\bottomrule
\end{tabular}
% }
\label{tab:LoRA}
\end{table}

\textbf{Comparisons on GEdit-Bench-CN.}
As shown in Table~\ref{tab:GEditCN}, we supplement the quantitative comparisons on GEdit-Bench-CN. HP-Edit still exhibits a obvious improvement across metrics, compared to Qwen-Image-Edit-2509, which demonstrates the effectiveness of our proposed framework. To quantify alignment, we conducted a new user study for HP-Edit on a held-out set from GEdit-Bench. As shown in Fig.~\ref{fig:validity}, human vs. HP-Scorer ratings show a strong concentration along the diagonal, yielding an average Pearson correlation coefficient (PCC) of \textbf{0.89}. Therefore, HP-Scorer is a valid evaluator and provides reliable reward signals for RL.

\begin{table}[h]
\centering
\caption{Performance of different methods on GEdit-Bench-CN.}
% \resizebox{\columnwidth}{!}{
%&7.647 &7.398 &6.983
\footnotesize
\begin{tabular}{l|lll}

\toprule
Model                      & \multicolumn{3}{l}{GEdit-Bench-CN ↑} \\ \cline{2-4} 
                                            & G\_SC       & G\_PQ      & G\_O      \\ \hline
Step1X-Edit~\cite{liu2025step1x}          &7.65 &7.40 &6.98     \\
X2Edit~\cite{ma2025x2edit}               &6.80             & 8.37           & 7.03        \\
% FLUX.1-Kontext-Dev~\cite{labs2025flux}   &     0.84    &    8.63    &   0.94    \\ \hline
Qwen-Image-Edit-2509~\cite{wu2025qwen} &   8.16      &    8.44    &   8.12    \\
HP-Edit              & \textbf{8.35}       &\textbf{8.54}     & \textbf{8.30}      \\ \bottomrule
\end{tabular}
% }
\label{tab:GEditCN}
\end{table}

\textbf{Comparisons on DreamBench++.}
As shown in Table~\ref{tab:DreamBench} and Table~\ref{tab:DreamBench2}, we compare the performance of HP-Edit with Qwen-Image-Edit-2509, and the results clearly demonstrate the improvement brought by HP-Edit.
\begin{table}[h]
    \centering
    \caption{Compare the performance of HP-Edit and the baseline model using traditional metrics on DreamBench++.}
    \resizebox{\columnwidth}{!}{
    \begin{tabular}{l|cccc}
    \toprule
    Methods     & DINO-I &CLIP-I & CLIP-T \\ \hline
    Qwen-Image-Edit-2509~\cite{wu2025qwen} & 0.504 & 0.749 & 0.346       \\
    HP-Edit    & \textbf{0.509} & \textbf{0.755} & \textbf{0.349}        \\
 \bottomrule
    \end{tabular}
    }
    \label{tab:DreamBench}
\end{table}

\begin{table*}[h]
    \centering
    \caption{Comparison of DreamBench++ results between HP-Edit and baseline, with scores for Concept Preservation (CP) and Prompt Following (PF).}
    \resizebox{\textwidth}{!}{
    \begin{tabular}{l|ccccc|cccc|cc}
    \toprule
    \multirow{2}{*}{Methods} & \multicolumn{5}{c|}{Concept Preservation} &  \multicolumn{4}{c|}{Prompt Following}  & \multirow{2}{*}{CP $\cdot$ PF} & \multirow{2}{*}{CP / PF} \\ \cline{2-10}
     & Animal & Human & Object & Style & Overall & Photorealistic & Style Transer & Imaginative & Overall & & \\
     \hline
    Qwen-Image-Edit-2509~\cite{wu2025qwen} & 0.646 & 0.509 & 0.663 & 0.509 & 0.617 & 0.946 & 0.935 & 0.900  & 0.932 & 0.575 & 0.662 \\
    HP-Edit  & \textbf{0.674} & \textbf{0.580} & \textbf{0.676} & \textbf{0.611} & \textbf{0.654} & \textbf{0.964} & \textbf{0.973} & \textbf{0.948} & \textbf{0.963} & \textbf{0.630}  & \textbf{0.679}   \\
    \bottomrule
    \end{tabular}
    }
    \label{tab:DreamBench2}
\end{table*}

% \begin{figure}[th] % 使用 [t] 或 [h] 提高紧凑度
%   \centering
% % \vspace{-2mm} % 压缩标题与正文之间的间距
%   \includegraphics[width=1.0\linewidth]{figure/pearson_correlation_square.pdf}
%   % \vspace{-4.5mm} % 压缩图片与标题之间的间距
%   \caption{Correlation analysis between human and HP-Scorer.}
%   \label{fig:validity}
%   % \vspace{-2mm} % 压缩标题与正文之间的间距
% \end{figure}

% \begin{figure*}[!t]
%   \resizebox{\columnwidth}{!}{
%   \includegraphics[width=\textwidth]{figure/pearson_correlation_square.pdf}}
%   \caption{Correlation analysis between human and HP-Scorer.}
%   \label{fig:validity}
% \end{figure*}

\textbf{Comparison with DPO.} We compare GRPO against DPO on the same subset ($>500$ cases/task, $5$ samples/case). DPO relies on offline winner/loser mining (often requiring repeated sampling and manual filtering), while GRPO performs online sampling with HP-Scorer feedback, which better explores the preference space. As shown below, DPO improves over the base model but remains worse than GRPO (HP-Scorer) and HP-Edit.

\begin{table}[th]
\caption{Comparison with DPO on RealPref-Bench}
 \resizebox{\linewidth}{!}{
\begin{tabular}{l|cccc}
\hline
         & base & w/DPO & w/ GRPO(HP-scorer) & HP-Edit\\ \hline
HP-score &  4.472    &  4.521     &     4.590                &       4.667            \\ \hline
\end{tabular}
 }
\end{table}

\section{More visual comparison}
\label{sec:visual}
We provide additional image editing results generated by HP-Edit in the following Figures~\ref{fig:spl2_remove},\ref{fig:spl2_add},\ref{fig:spl2_swap},\ref{fig:spl2_bgreplace},\ref{fig:spl2_bokeh}, \ref{fig:spl2_relight},\ref{fig:spl2_style},\ref{fig:spl2_color}.

\begin{figure*}[!t]
    \centering
    \includegraphics[width=0.9\textwidth]{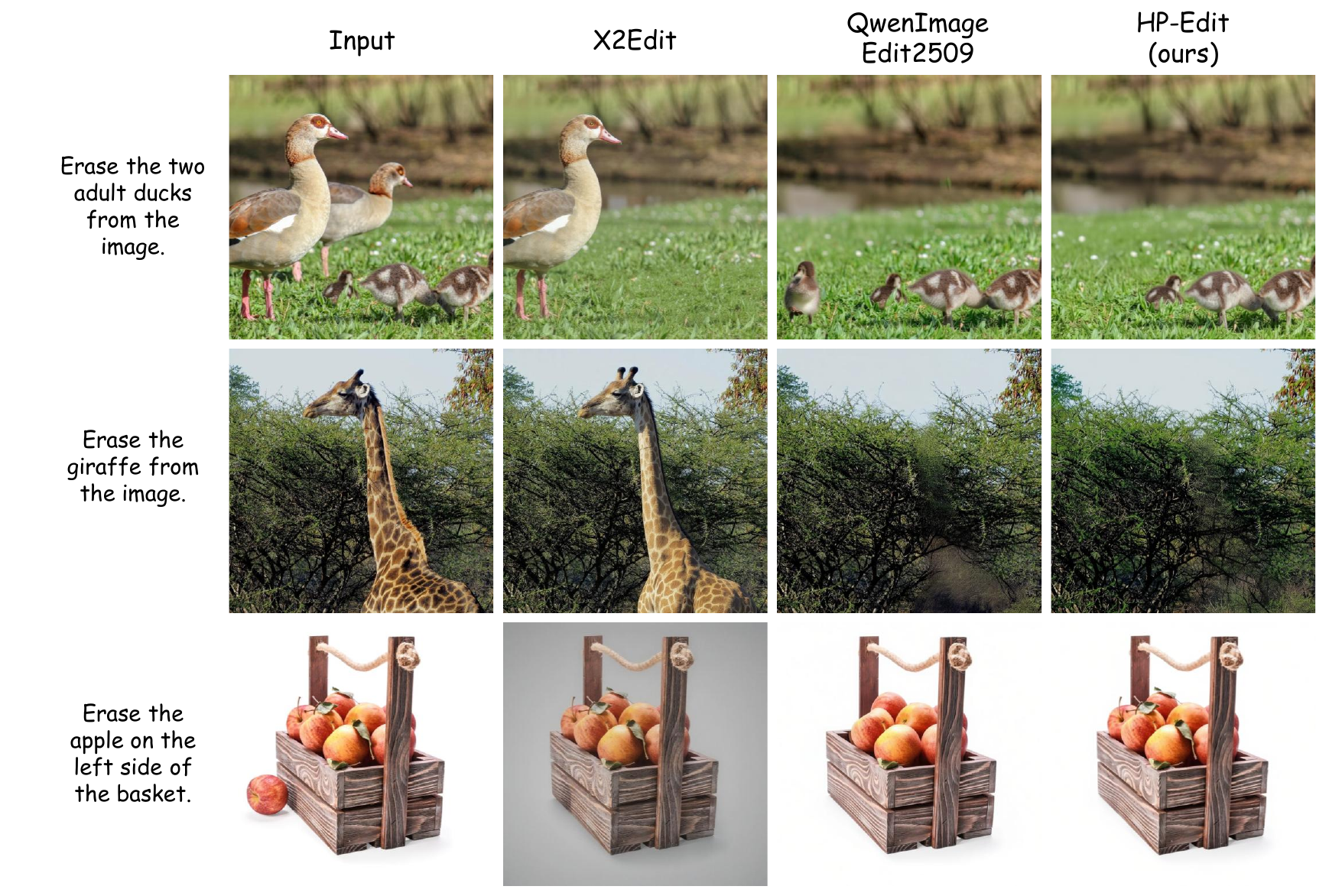} 
    \caption{Qualitative comparison of object removal task.} 
    \vspace{-0.3cm}
    \label{fig:spl2_remove}
\end{figure*}

\begin{figure*}[!t]
    \centering
    \includegraphics[width=0.9\textwidth]{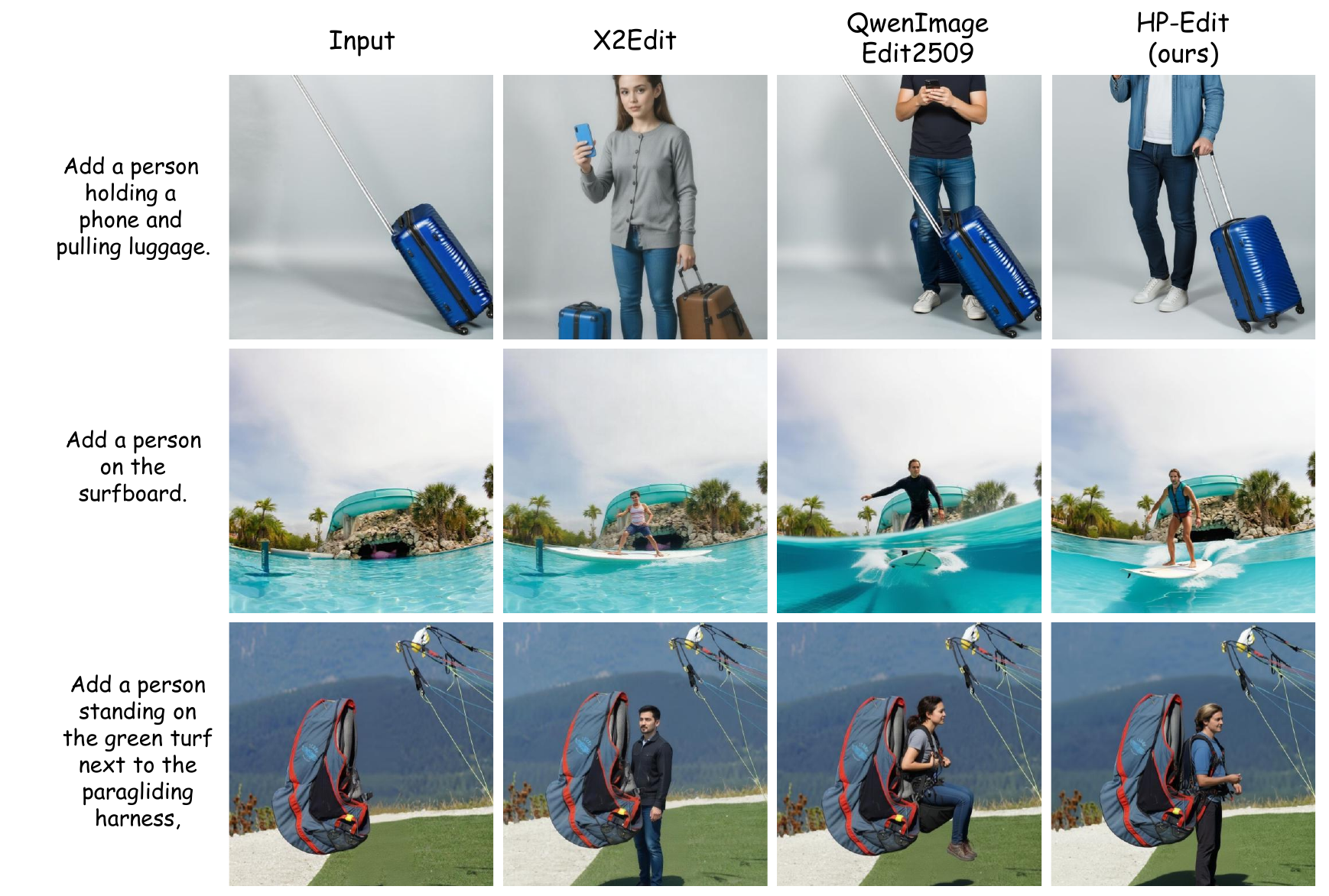} 
    \caption{Qualitative comparison of object adding task.} 
    \vspace{-0.3cm}
    \label{fig:spl2_add}
\end{figure*}

\begin{figure*}[!t]
    \centering
    \includegraphics[width=0.9\textwidth]{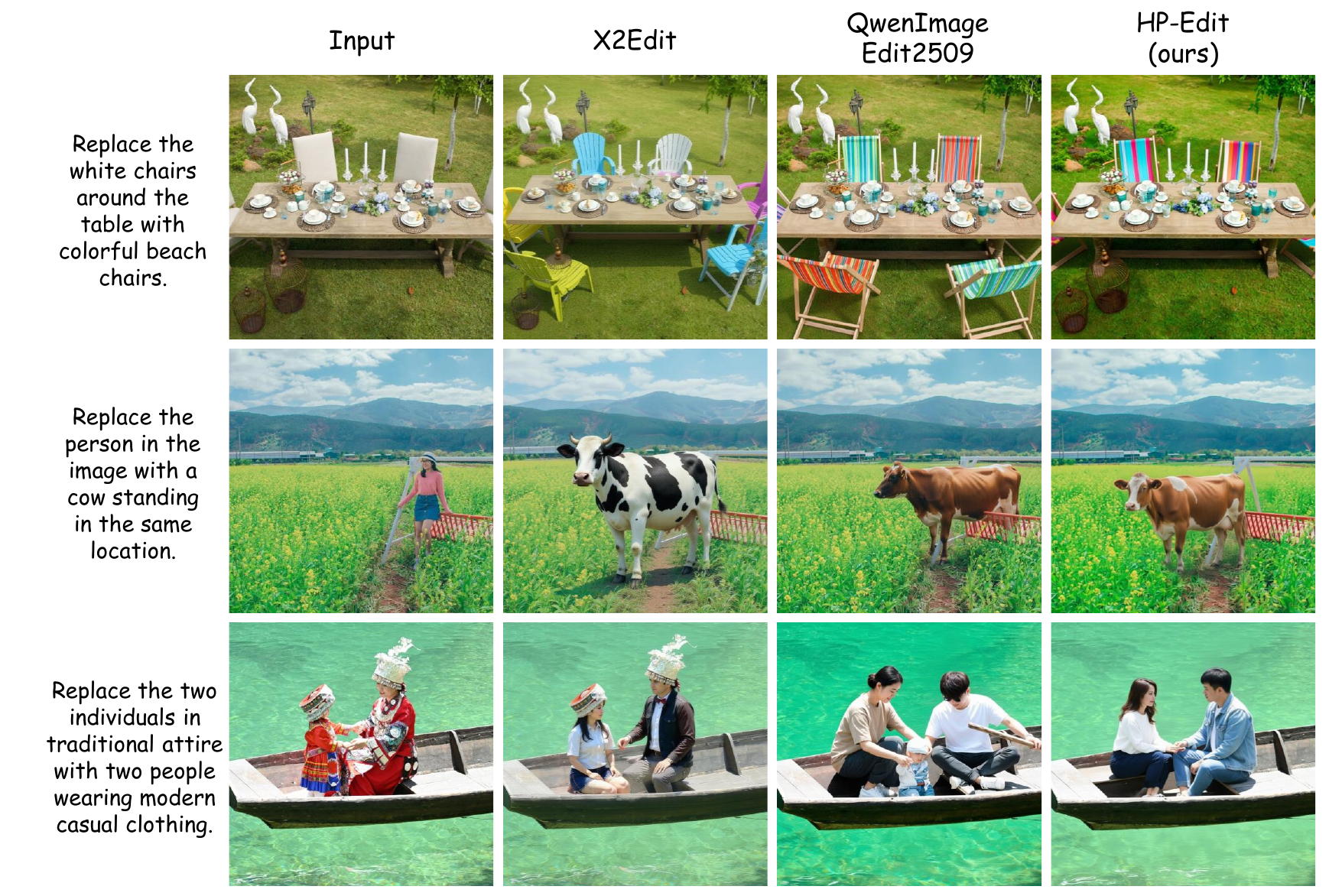} 
    \caption{Qualitative comparison of object swapping task.} 
    \vspace{-0.3cm}
    \label{fig:spl2_swap}
\end{figure*}

\begin{figure*}[!t]
    \centering
    \includegraphics[width=0.9\textwidth]{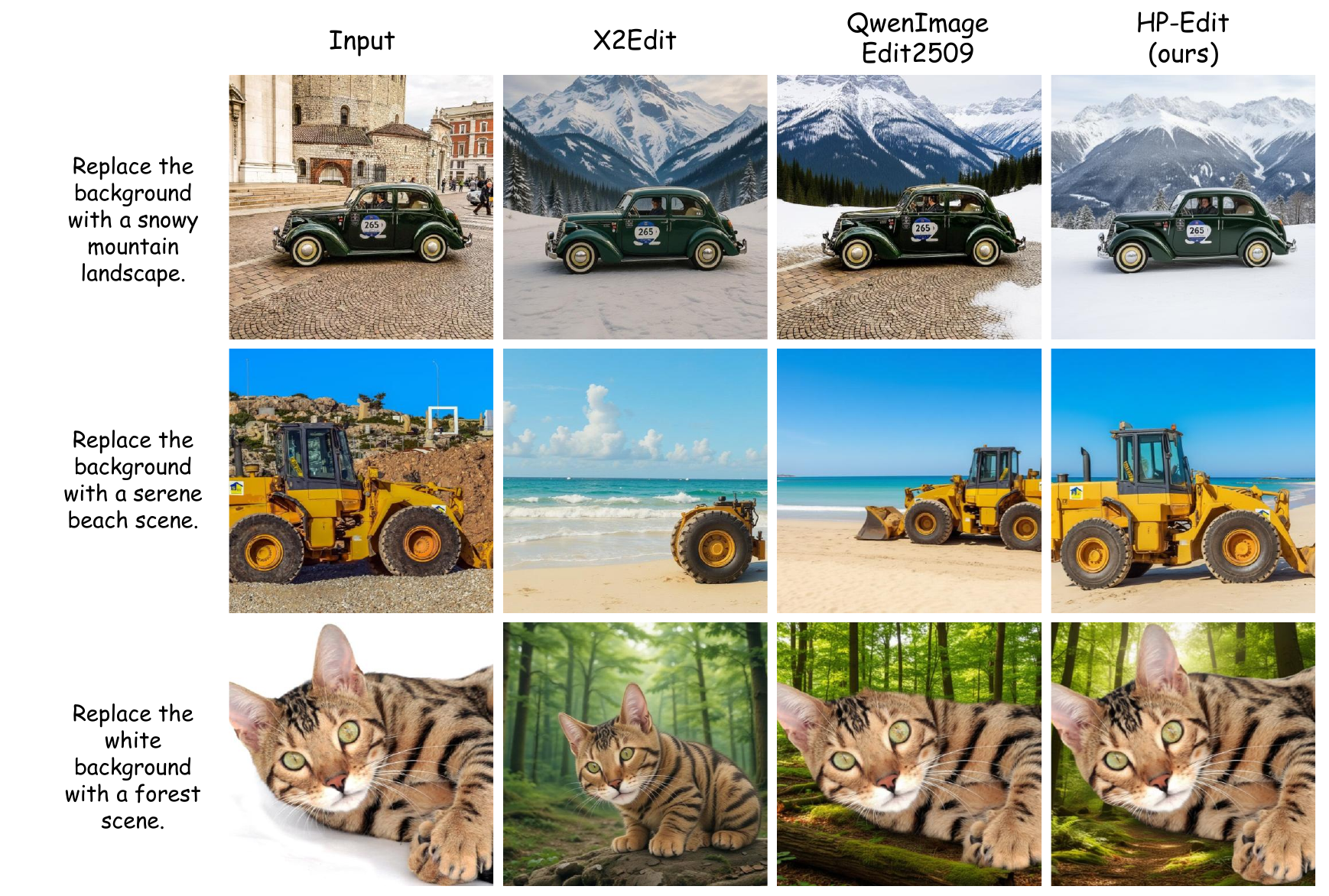} 
    \caption{Qualitative comparison of background replacement task.} 
    \vspace{-0.3cm}
    \label{fig:spl2_bgreplace}
\end{figure*}

\begin{figure*}[!t]
    \centering
    \includegraphics[width=0.9\textwidth]{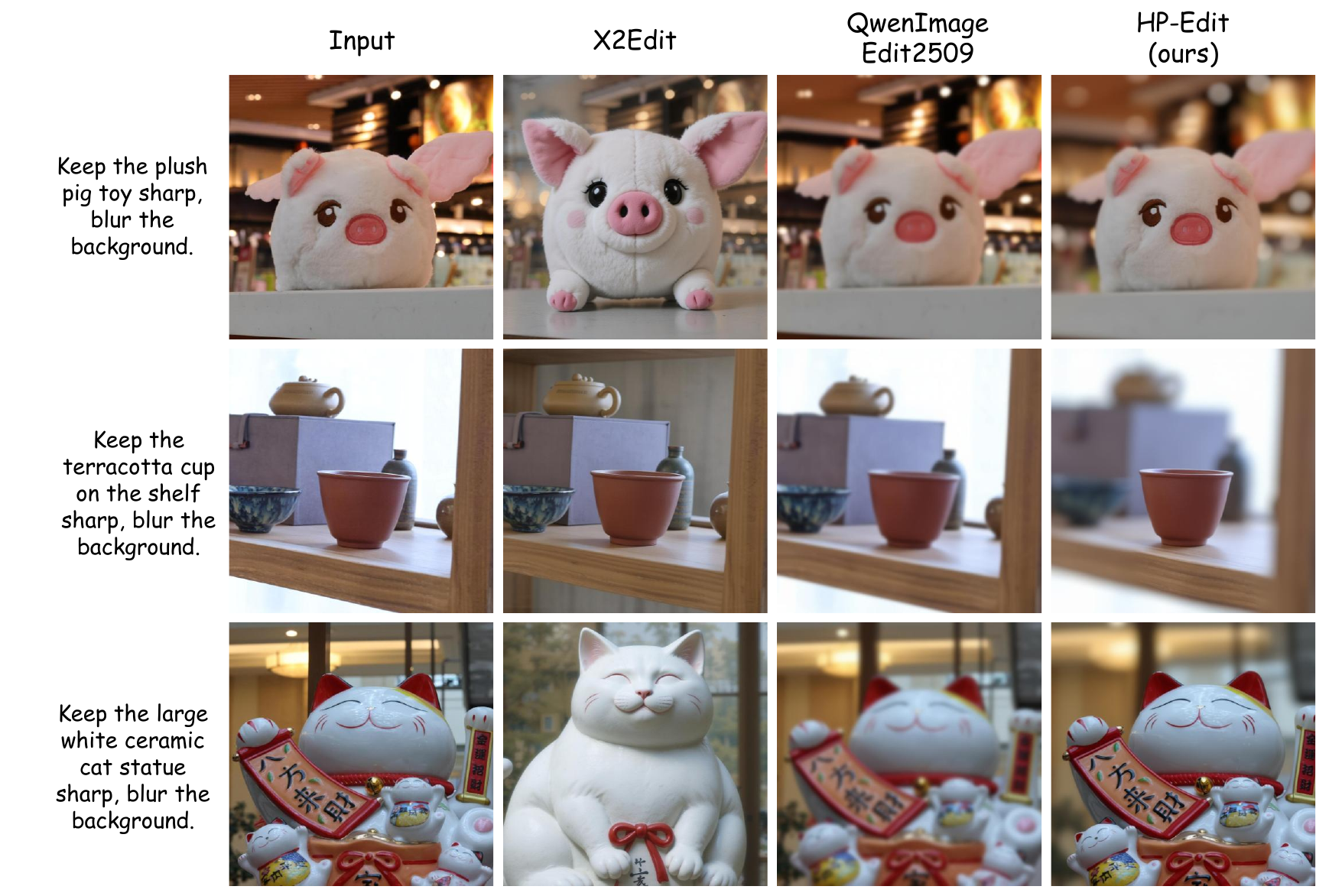} 
    \caption{Qualitative comparison of bokeh task.} 
    \vspace{-0.3cm}
    \label{fig:spl2_bokeh}
\end{figure*}

\begin{figure*}[!t]
    \centering
    \includegraphics[width=0.9\textwidth]{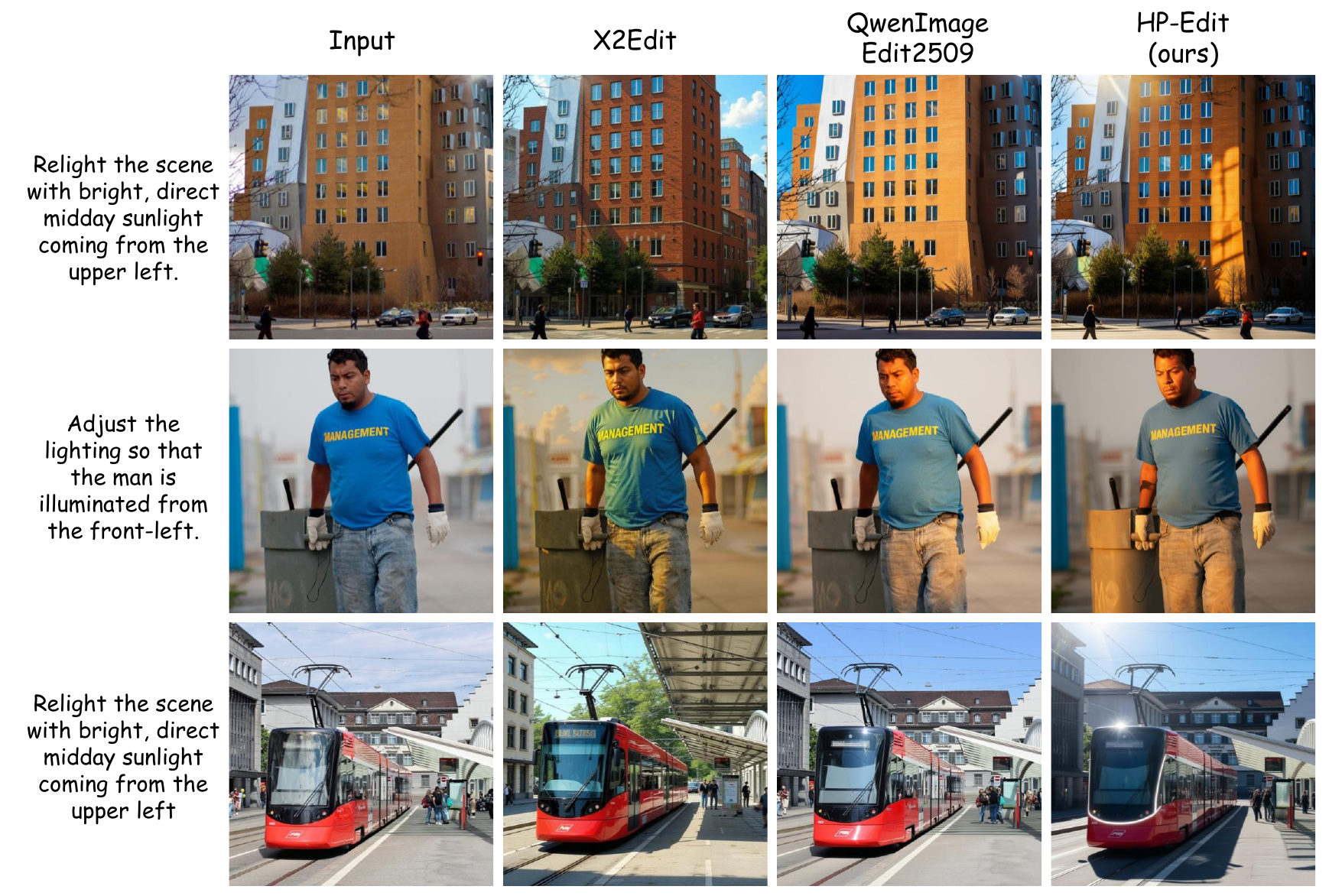} 
    \caption{Qualitative comparison of relighting task.} 
    \vspace{-0.3cm}
    \label{fig:spl2_relight}
\end{figure*}

\begin{figure*}[!t]
    \centering
    \includegraphics[width=0.9\textwidth]{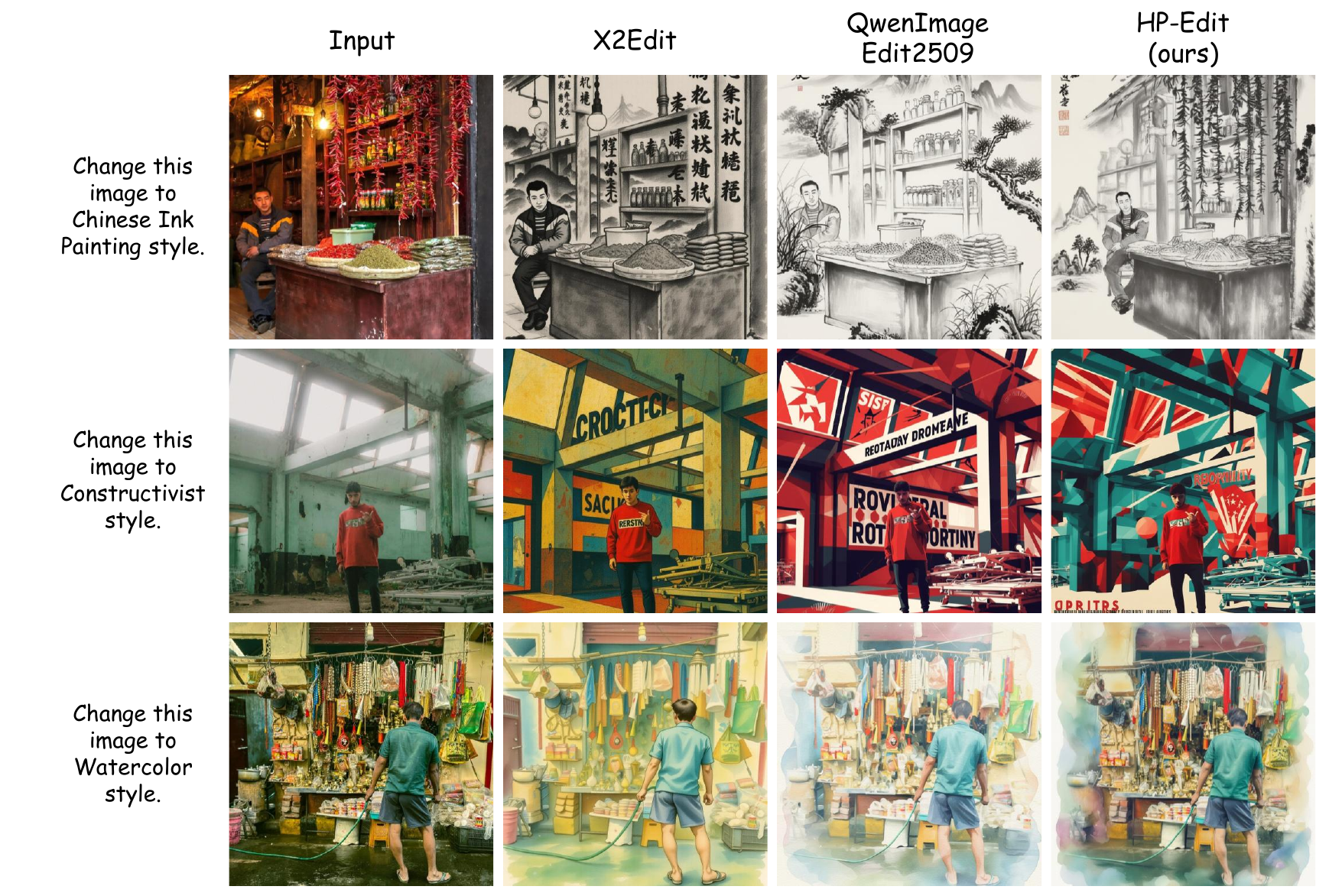} 
    \caption{Qualitative comparison of style changing task.} 
    \vspace{-0.3cm}
    \label{fig:spl2_style}
\end{figure*}

\begin{figure*}[!t]
    \centering
    \includegraphics[width=0.9\textwidth]{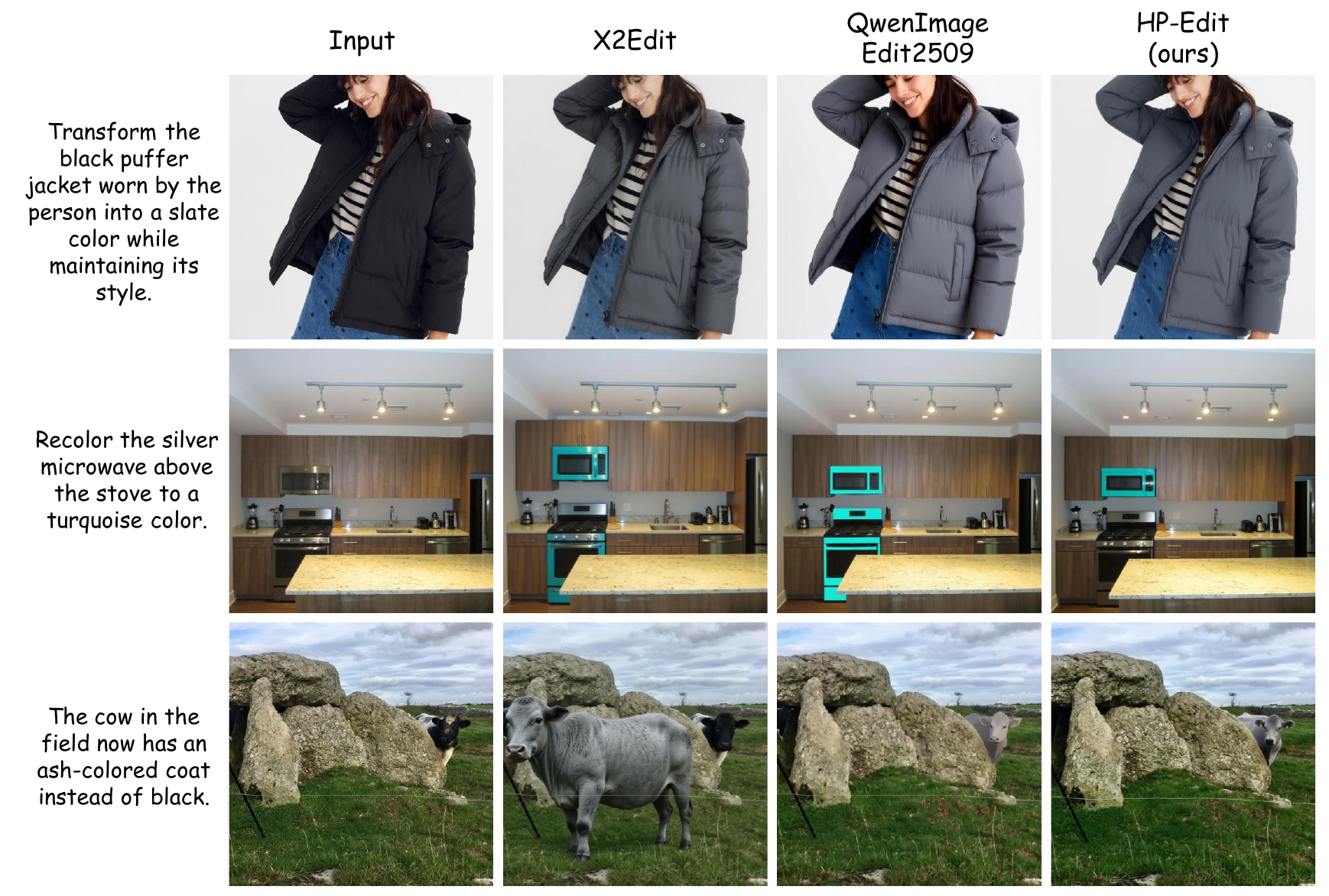} 
    \caption{Qualitative comparison of color changing task.} 
    \vspace{-0.3cm}
    \label{fig:spl2_color}
\end{figure*}

\section{Details of RealPref-50k and RealPref-Bench}
\label{sec:data}
Table~\ref{tab:task_stats} presents the statistics of RealPref-50K and RealPref-Bench across eight editing tasks. 

% In addition, all editing instructions in the RealPref-Bench are manually validated to ensure the instruction quality and editing fidelty.

\begin{table}[th]
\centering
\caption{Task statistics across RealPref-bench and Realpref-50K.}
\label{tab:task_stats}
\resizebox{\columnwidth}{!}{% % Ensures the table fits the column width
\begin{tabular}{lcccr}
\toprule
\textbf{Task Name} & \textbf{Benchmark Count} & \textbf{Dataset Ratio} & \textbf{Dataset Count} \\
\textbf{} & \textbf{(Realpref-bench)} & \textbf{(\% of Realpref-50K)} & \textbf{(Realpref-50K)} \\
\midrule
Add & 232 & 14.16\% & 7,903 \\
Background Replace & 191 & 11.66\% & 6,506 \\
Bokeh & 200 & 12.21\% & 6,813 \\
Color& 204 & 12.45\% & 6,949 \\
Object Swap & 242 & 14.77\% & 8,243 \\
Relight & 150 & 9.16\% & 5,109 \\
Removal & 227 & 13.86\% & 7,732 \\
Stylize & 192 & 11.72\% & 6,540 \\
\midrule
\textbf{Total / Sum} & \textbf{1,638} & \textbf{100.00\%} & \textbf{55,795} \\
\bottomrule
\end{tabular}
}
\end{table}

To illustrate the dataset construction process, we highlight two representative tasks: style transfer and bokeh. For the style transfer task, we first collect content images from high-resolution datasets (e.g., Div2K) and real-world photo collections. Style references span more than 30 categories, including classical artistic styles (e.g., Impressionism, ink-wash painting) and contemporary aesthetics (e.g., anime), with 20–30 examples per category. The editing instructions explicitly enforce the target style while preserving the original structure (e.g., “convert this image into LEGO style while maintaining object layout and geometry”).

For the bokeh task, we collect aligned bokeh–non-bokeh image pairs from existing datasets. A pretrained VLM is then used to generate region-specific editing instructions that emphasize depth-of-field changes in the focused regions. The images are balanced according to COCO object classes.

Similarly, the remaining editing tasks (e.g., object swapping, object removal, background replacement, attribute modification, relighting, and composition editing) are constructed by combining high-quality image sources, VLM-generated editing instructions, and task-specific filtering rules to ensure diversity and realism.

\end{document}